\documentclass[%
 reprint,
 amsmath,amssymb,
 aps,
]{revtex4-2}

\usepackage{color}
\usepackage{comment}
\usepackage{graphicx}
\usepackage{dcolumn}
\usepackage{bm}
\usepackage{algorithm}
\usepackage{algorithmic}
\usepackage{bbold}
\usepackage{soul}
\usepackage{chngcntr}
\usepackage{bm}
\newtheorem{prop}{Proposition}
\numberwithin{prop}{subsection}
\newtheorem{coro}{Corollary}
\numberwithin{coro}{subsection}
\newtheorem{defn}{Definition}
\usepackage{lipsum}
\usepackage{hyperref}

\begin{document}

\preprint{APS/123-QED}

\title{Inverse-Dirichlet Weighting Enables Reliable Training of Physics Informed Neural Networks}

\author{Suryanarayana Maddu$^{1,2,3,7}$, Dominik Sturm$^{9,10}$, Christian L.~M\"{u}ller$^{4,5,6}$, Ivo F.~Sbalzarini$^{1,2,3,7,8}$} \email{ivos@mpi-cbg.de}
\affiliation{ $^{1}$ Technische Universit\"{a}t Dresden, Faculty of Computer Science, 01069 Dresden, Germany}
\affiliation{$^{2}$ Max Planck Institute of Molecular Cell Biology and Genetics, 01307 Dresden, Germany}
\affiliation{ $^{3}$ Center for Systems Biology Dresden, 01307 Dresden, Germany} 
\affiliation{ $^{4}$ Center for Computational Mathematics, Flatiron Institute, New York, NY, USA}
\affiliation{ $^{5}$ Department of Statistics, LMU M\"{u}nchen, Munich, Germany}
\affiliation{ $^{6}$ Institute of Computational Biology, Helmholtz Zentrum M\"{u}nchen, Germany}
\affiliation{ $^{7}$ Center for Scalable Data Analytics and Artificial Intelligence ScaDS.AI, Dresden/Leipzig, Germany}
\affiliation{ $^{8}$ Cluster of Excellence Physics of Life, TU Dresden, Germany}
\affiliation{ $^{9}$ Helmholtz-Zentrum Dresden-Rossendorf, D-01328 Dresden, Germany}
\affiliation{ $^{10}$ Center for Advanced Systems Understanding (CASUS), D-02826 G\"{o}rlitz, Germany}

\date{\today}

\begin{abstract}
We characterize and remedy a failure mode that may arise from multi-scale dynamics with scale imbalances during training of deep neural networks, such as Physics Informed Neural Networks (PINNs). PINNs are popular machine-learning templates that allow for seamless integration of physical equation models with data.
Their training amounts to solving an optimization problem over a weighted sum of data-fidelity and equation-fidelity objectives. Conflicts between objectives can arise from scale imbalances, heteroscedasticity in the data, stiffness of the physical equation, or from catastrophic interference during sequential training. 
We explain the training pathology arising from this and propose a simple yet effective inverse-Dirichlet weighting strategy to alleviate the issue.
We compare with Sobolev training of neural networks, providing the baseline of analytically $\boldsymbol{\epsilon}$-optimal training.
We demonstrate the effectiveness of inverse-Dirichlet weighting in various applications, including a multi-scale model of active turbulence, where we show orders of magnitude improvement in accuracy and convergence over conventional PINN training. For inverse modeling using sequential  training, we find that inverse-Dirichlet weighting protects a PINN against catastrophic forgetting.
\end{abstract}

\maketitle


\section{Introduction}

Data-driven modeling has emerged as a powerful and complementary approach to first-principles modeling. It was made possible by  advances in imaging and measurement technology, computing power, and  innovations in machine learning, in particular neural networks. The success of neural networks in data-driven modeling can be attributed to their powerful function approximation properties for diverse functional forms, from image recognition \cite{he2016deep} and natural language processing \cite{vanwavenet} to learning policies in complex environments \cite{silver2016mastering}. 

Recently, there has been a surge in exploiting the versatile approximation properties of neural networks for surrogate modeling in scientific computing and for data-driven modeling of physical systems~\cite{sirignano2018dgm, raissi2019physics}. These applications hinge on the approximation properties of neural networks for Sobolev-regular functions \cite{guhring2020error, czarnecki2017sobolev}. The popular Physics Informed Neural Networks (PINNs) rely on knowing a differential equation model of the system in order to solve a soft-constrained optimization problem \cite{raissi2019deepvortex, raissi2019physics}. Thanks to their mesh-free character, PINNs can be used for both forward and inverse modeling in domains including material science \cite{fang2019deep, liu2019multi, chen2020physics}, fluid dynamics \cite{raissi2019deepvortex,raissi2020hidden, sun2020surrogate, mao2020physics} and turbulence \cite{raissi2019deep, jin2020nsfnets}, biology \cite{yazdani2020systems}, medicine \cite{sahli2020physics, kissas2020machine}, earth science \cite{zheng2020physics}, mechanics \cite{rao2021physics}, uncertainty quantification \cite{yang2020b}, as well as stochastic \cite{zhang2020learning}, high-dimensional  \cite{han2018solving}, and fractional differential equations \cite{pang2019fpinns}.

The wide application scope of PINNs rests on their parameterization as deep neural networks, which can be tuned by minimizing a weighted sum of data-fitting and equation-fitting errors. However, finding network parameters that impartially optimize for several errors is challenging when errors impose conflicting requirements. For Multi-Task Learning (MTL) problems in machine learning, this issue has been addressed by various strategies for relative weighting of conflicting objectives, e.g., based on uncertainty \cite{kendall2018multi}, gradient normalization \cite{chen2018gradnorm}, or Pareto optimality \cite{sener2018multi}. The PINN community in parallel has developed heuristics for  loss weighting with demonstrated gains in accuracy and convergence \cite{wang2020understanding}. However, there is neither consensus on when to use such strategies in PINNs, nor are there optimal benchmark solutions for objectives based on differential equations. 

Here, we characterize training pathologies of PINNs and provide criteria for their occurrence. We explain these pathologies by showing a connection between PINNs and Sobolev training, and we propose a strategy for loss weighting in PINNs based on the Dirichlet energy of the task-specific gradients. We show that this strategy reduces optimization bias and protects against catastrophic forgetting. We evaluate the proposed inverse-Dirichlet weighting by comparing with Sobolev training in a case where provably optimal weights can be derived, and by empirical comparison with two conceptually different state-of-the-art PINN weighting approaches.

\section{A Physics Informed Neural Network is a Multi-Objective Optimization Problem}
An optimization problem involving multiple tasks or objectives can be written as:
\begin{equation}\label{eq:multiobj}
    \textrm{minimize  } \mathcal{L}_k(\bm{\theta}^{sh}, \bm{\theta}^k), \qquad k = 1,\ldots ,K.
\end{equation}
The total number of tasks is given by $K$. $\bm{\theta}^{sh}$ are shared parameters between the tasks, and $\bm{\theta}^k$ are the task-specific parameters. For conflicting objectives, there is no single optimal solution, and Pareto dominated solutions can be found using heuristics like evolutionary algorithms~\cite{zitzler1999multiobjective}. Alternatively, the multi-objective optimization problem can be converted to a single-objective optimization problem using scalarization, e.g., as a weighted sum with real-valued relative weights $\lambda_k$ associated with each loss objective $\mathcal{L}_k$:
\begin{equation}
    \min_{\bm{\theta}^{sh}, \bm{\theta}^1,\ldots ,\bm{\theta}^K}  \sum_{k=1}^{K}\lambda_k \mathcal{L}_{k} (\bm{\theta}^{sh}, \bm{\theta}^k).
\end{equation}
Often, the $\lambda_k$ are dynamically tuned along training epochs~\cite{sener2018multi}. 
The weights $\lambda_k$ intuitively quantify the extent to which we aim to minimize the objective $\mathcal{L}_k$. Large $\lambda_k$ favor small $\mathcal{L}_k$, and vice-versa~\cite{boyd2004convex}. For $K \geq 2$, manual tuning and grid search become prohibitively expensive for large parameterizations as common in deep neural networks. 

The loss function of a PINN is a scalarized multi-objective loss where each objective corresponds to either a data-fidelity term or a constraint derived from the physics of the underlying problem~\cite{raissi2019physics, sirignano2018dgm}.
A typical PINN loss contains five objectives:
\begin{multline}\label{eq:PINN}
    \mathcal{L}(\bm{u}_{\bm{\theta}}, \bm{u}) = \frac{\lambda_\text{1}}{N_\text{int}} \sum_{i,j=1}^{N_\text{int}} \Big \vert \partial_t \bm{u}_{\bm{\theta}} - \mathcal{N} (\bm{u}_{\bm{\theta}}(t_j, \bm{x}_i^{\Omega}), \Sigma )  \Big \vert^2 + \\
     \frac{\lambda_2}{N_\mathcal{B}} \sum_{i,j=1}^{N_\mathcal{B}} \Big \vert \mathcal{B} (t_j, \bm{x}_i^{\partial \Omega}) \Big \vert^2
    +  \frac{\lambda_3}{N_\text{I}} \sum_{i=1}^{N_\text{I}} \Big \vert \mathcal{I} (t =0 , \bm{x}_i^{\Omega}) \Big \vert^2 +\\
     \frac{\lambda_4}{N_\text{int}} \sum_{i,j=1}^{N_\text{int}} \Big \vert \mathcal{F} (\bm{u}_{\bm{\theta}}(t_j, \bm{x}_i^{\Omega}) ) \Big \vert^2 +  \frac{\lambda_5}{N_\text{int}} \sum_{i,j=1}^{N_\text{int}} \Big \vert \bm{u}_{\bm{\theta}}(t_j, \bm{x}_i^{\Omega}) - \bm{u}_i ) \Big \vert^2 .
\end{multline}
We distinguish the neural network approximation $\bm{u}_{\bm{\theta}}(t_j,\bm{x}_i^\Omega)$ from the training data $\bm{u}_i = \bm{u}(t_j, \bm{x}_i^\Omega)$ sampled at the $N_\text{int}$ space-time points $(t_j,\bm{x}_i^{\Omega})$ in the interior of the solution domain $\Omega$, $N_\mathcal{B}$ points  $(t_j,\bm{x}_i^{\partial \Omega})$ on the boundary $\partial \Omega$, and $N_\text{I}$ samples of the initial state at time $t=0$. The $\lambda_1$, $\lambda_2$, and $\lambda_3$ weight the residuals of the physical equation $\partial_t \bm{u} = \mathcal{N} (\bm{u}, \Sigma )$ with right-hand side $\mathcal{N}$ and coefficients $\Sigma$, the boundary condition $\mathcal{B}$, and the initial condition $\mathcal{I}$, respectively. The objective with weight $\lambda_4$ is an additional constraint $\mathcal{F}$ on the state variable $\bm{u}$, such as divergence-freeness ($\nabla \cdot \bm{u} = 0$) of the velocity field of an incompressible flow. The weight $\lambda_5$ penalizes deviations of the neural network approximation from the training data. From now on, we simplify the notation by designating space-time points as $\bm{x}_i := (t,\bm{x})_i = (t_j,\bm{x}_k)$ in time-dependent problems.

PINNs can be used to solve both forward and inverse modeling problems of ordinary differential equations (ODEs) or partial differential equations (PDEs). In the forward problem, a numerical approximation to the solution of a ODE or PDE is to be learned. For this, $\bm{\theta}^\text{sh}$ are the networks parameters, and $\bm{\theta}^{k}$ is an empty set. In the inverse problem, the coefficients of an ODE or PDE are to be learned from data such that the equation describes the dynamics in the data. Then, the task-specific parameters $\bm{\theta}^{k}$ correspond to the coefficients ($\bm{\xi} \subseteq \Sigma $) of the physical equation that shall be inferred. During training of a PINN, the parameters $(\bm{\theta}^\text{sh}, \bm{\theta}^{k})$ are determined by minimizing the total loss $\mathcal{L}$ for given data $\bm{u}(\bm{x}_i)$.

\subsection{Sobolev training is a special case of PINN training}
Like most deep neural networks, PINNs are usually trained by first-order parameter update based on (approximate) loss gradients. The update rule can be interpreted as a discretization of the gradient flow, i.e.,
\begin{align}\label{eq:updaterule}
\bm{\theta} (\tau + 1) &= \bm{\theta} (\tau) - \eta (\tau)  \sum_{k=1}^{K}\lambda_k  \nabla_{\bm{\theta}} \mathcal{L}_{k}.
\end{align}
The learning rate $\eta(\tau)$ depends on the training epoch $\tau$, reflecting the adaptive step size used in optimizers like Adam \cite{DBLP:journals/corr/KingmaB14}. The gradients are batch gradients $ \nabla_{\bm{\theta}} \mathcal{L}_k = \frac{1}{\vert \textrm{B} \vert} \sum_{i \in \textrm{B}} \nabla \ell_k (\bm{u}_{\bm{\theta}}(\bm{x}_i), \bm{u}_i) $,
 where $\textrm{B}$ is a mini-batch created by random splitting of the $N$ training data points into $N/\vert \textrm{B}\vert$ distinct partitions and $\ell_k$ is one summand (for one data point $\bm{x}_i$) of the loss $\mathcal{L}_k$ corresponding to the $k^{\text{th}}$ objective. For batch size $\vert \textrm{B}\vert=1$, the mini-batch gradient descent reduces to stochastic gradient descent.

Adding derivatives $\mathcal{D}_{\textbf{x}}^m$, $m\geq 1$, of the state variable into the training loss has been shown to improve data efficiency and generalization power of neural networks~\cite{czarnecki2017sobolev}. Minimizing the resulting loss 
 \begin{equation}\label{eq:Sobolev}
  \sum_{i=1}^{N} \Bigg[ \lambda_0 \vert  \bm{u}_{\bm{\theta}} (\bm{x}_i) - \bm{u}(\bm{x}_i) \vert^2  + \sum_{k=1}^{K}  \lambda_k \vert \mathcal{D}_{\textbf{x}}^k  \bm{u}_{\bm{\theta}} (\bm{x}_i) -  \mathcal{D}_{\textbf{x}}^k \bm{u}(\bm{x}_i) \vert^2 \Bigg ]
\end{equation}
using the update rule in Eq.~\ref{eq:updaterule} can be interpreted as a finite approximation to the Sobolev gradient flow of the loss functional.
The neural network is then trained with access to the ($K+2$)-tuples $\{ (\bm{x}_i, \bm{u}(\bm{x}_i), \mathcal{D}_{\textbf{x}}^1,\ldots,\mathcal{D}_{\textbf{x}}^K ) \}_{i=1}^N$ instead of the usual training set $\{ (\bm{x}_i, \bm{u}(\bm{x}_i))\}_{i=1}^N$. Such Sobolev training is an instance of scalarized multi-objective optimization with each objective based on the approximation of a specific derivative. The weights are usually chosen as $\lambda_0 = \ldots =\lambda_\text{K}=1$~\cite{czarnecki2017sobolev}. The additional information from the  derivatives introduces an inductive bias, which has been found to improve data efficiency and network generalization~\cite{czarnecki2017sobolev}. This is akin to PINNs, which use derivatives from the PDE model to achieve inductive bias. Indeed, the Sobolev loss in Eq.~\ref{eq:Sobolev} is a special case of the PINN loss from Eq.~\ref{eq:PINN} with each objective based on a PDE with right-hand side $\mathcal{N} =  \mathcal{D}_{\textbf{x}}^k \bm{u}$, $k = 1,\ldots ,K$. 

\subsection{Vanishing task-specific gradients are a failure mode of PINNs}
The update rule in Eq.~\ref{eq:updaterule} is known to lead to stiff learning dynamics for loss functionals whose Hessian matrix has large positive eigenvalues \cite{wang2020understanding}. 
This is the case for Sobolev training, even with $K=1$, when the function $\bm{u}(\bm{x})$ is highly oscillatory, leading us to the following proposition, proven in the Supplementary Material:
\begin{prop}\label{prop}
[{\textbf{Stiff learning dynamics in first-order parameter update}}] For Sobolev training on the tuples $\{ \bm{x}_i, \bm{u}(\bm{x}_i), \partial_{\bm{x}}^m \bm{u}(\bm{x}_i) \}_{i=1}^{N}, m \geq 1$, with total loss $\mathcal{L} =  \lambda_1 \sum_{i=1}^{N}\vert \bm{u}_{\bm{\theta}}(\bm{x}_i) - \bm{u}(\bm{x}_i) \vert^2 + \lambda_2 \vert \mathcal{D}_{\bm{x}}^m \bm{u}_{\bm{\theta}} (\bm{x}_i) - \mathcal{D}_{\bm{x}}^m \bm{u}(\bm{x}_i)\vert^2$, the rate of change of the residual of the dominant Fourier mode $\widetilde{\bm{r}}(\bm{k}_0) = \widetilde{\bm{u}}_{\bm{\theta}} (\bm{k}_0) - \widetilde{\bm{u}} (\bm{k}_0)$ using first-order parameter update is:
\begin{align*}
   \Big \vert \frac{\mathrm{d}\widetilde{\bm{r}}({\bm{k}_0})}{\mathrm{d}\tau} \Big \vert & = \frac{\eta}{O(\bm{k}_0) } \Big[ \lambda_1 O(1)  +\lambda_2 O(\bm{k}_0^{2m})  \Big ] \Big \vert \frac{\partial \widetilde{\bm{r}}({\bm{k}_0})}{\partial \bm{\theta}} \Big \vert ,
\end{align*}
where $O(\cdot)$ is the Bachmann-Landau big-O symbol.
For $\lambda_1 = \lambda_2 = 1$, this leads to training dynamics with slow time scale $\mathrm{d}/\mathrm{d}t \in O(\bm{k}_0^{-1})$ and fast time scale $\mathrm{d}/\mathrm{d}t \in O(\bm{k}_0^{2m-1})$ for $\widetilde{\bm{r}}({\bm{k}_0}) \in O(1)$ ( Proof in supplementary ).
\end{prop}

This proposition suggests that losses containing high derivatives $m$, dominant high frequencies $\bm{k}_0$, or a combination of both exhibit dominant gradient statistics. This can lead to biased optimization of the dominant objectives at the expense of the other objectives with relatively smaller gradients. This is akin to the well known vanishing gradients phenomenon across the layers of a neural network \cite{glorot2010understanding}, albeit now across different objectives of a multi-objective optimization problem. We call this phenomenon \emph{vanishing task-specific gradients}.

Proposition \ref{prop} also admits a qualitative comparison to the phenomenon of numerical stiffness in PDEs of the form
\begin{equation}
    \frac{\partial {\widetilde{\bm{u}}}}{\partial t} = \mathbf{L} \widetilde{\bm{u}} + \mathbf{F}(\widetilde{\bm{u}}, t) 
\end{equation}
If the real parts of the eigenvalues of $\mathbf{L}$, $c = \mathrm{Re}(\lambda(\mathbf{L})) \gg 1$, then the solution  of the above PDE has fast modes $\widetilde{\bm{u}} \in O(1)$ with $\mathrm{d}/\mathrm{d}t \in O(c)$ and slow modes  $\widetilde{\bm{u}} \in O(1/c)$ with $\mathrm{d}/\mathrm{d}t \in O(1)$. In words, high frequency modes evolve on shorter time scales than low frequency modes. The rate amplitude for spatial derivatives of order $m$ is $\mathrm{d}/\mathrm{d}t \in O(\bm{k}^{-m})$ for the wavenumber $\bm{k}$~\cite{cox2002exponential}. Due to this analogy, we call the learning dynamics of a neural network with vanishing task-specific gradients \emph{stiff}. Stiff learning dynamics leads to discrepancy in the convergence rates of different objectives in a loss function.

Taken together, vanishing task-specific gradients constitute a likely failure mode of PINNs, since their training amounts to an extended version of Sobolev training.  

\section{Strategies for Balanced PINN Training}
Since PINN training minimizes a weighted sum of multiple objectives (Eq.~\ref{eq:PINN}), the result depends on appropriately chosen weights. 
The ratio between any two weights $\lambda_i/\lambda_j$, $i \neq j$, defines the relative importance of the two objectives during training. Suitable weighting is therefore quintessential to finding a Pareto-optimal solution that balances all objectives.
While \textit{ad hoc} manual tuning is still commonplace, it has recently been proposed to determine the weights as moving averages over the inverse gradient magnitude statistics of a given objective as \cite{wang2020understanding}:
\begin{align} \label{eq:max_avg}
    \hat{\lambda}_k (\tau) &= \frac{\max \{ \vert \nabla_{\bm{\theta}^{sh}} \mathcal{L}_\text{1} (\tau) \vert \} }{ \lambda_k (\tau) \overline{  \vert  \nabla_{\bm{\theta}^{sh}}  \mathcal{L}_k  (\tau)  \vert}},  \\
    \lambda_k (\tau + 1) &= \alpha \lambda_k(\tau) + (1-\alpha) \hat{\lambda}_k(\tau), \nonumber
\end{align}
where $\nabla_{\bm{\theta}^{sh}} \mathcal{L}_\text{1}$ is the gradient of the residual, i.e.~of the first objective in Eq.~\ref{eq:PINN}, and $\alpha$ is a user-defined learning rate (here always $\alpha=0.5$, see Supplementary Material). 
All gradients are taken with respect to the shared parameters across tasks and $| \cdot |$ is the component-wise absolute value. The overbar signifies the algebraic mean over the vector components. The maximum in the numerator is over all components of the gradient vector. 
While this weighting has been shown to improve PINN performance \cite{wang2020understanding}, it does not prevent vanishing task-specific gradients.

\subsection{Inverse-Dirichlet weighting} Instead of determining the relative weights proportional to the inverse average gradient magnitude, we here propose to use weights based on the gradient variance. In particular, we propose to use weights for which the variances over the components of the back-propagated weighted gradients $\lambda_k \nabla_{\bm{\theta}} \mathcal{L}_k$ become equal across all objectives, thus directly preventing vanishing task-specific  gradients. This can be achieved in any (stochastic) gradient descent optimizer by using the update rule in Eq.~\ref{eq:updaterule} with weights 
\begin{align}\label{eq:varnorm}
    \hat{\lambda}_k (\tau) &=  \frac{ \max_{t=1\ldots K} \left( \text{Var}[ \nabla_{\bm{\theta}^{sh}} \mathcal{L}_t (\tau) ] \right) }{\text{Var}[ \nabla_{\bm{\theta}^{sh}} \mathcal{L}_k (\tau) ]},   \\
        \lambda_k (\tau + 1) &= \alpha \lambda_k(\tau) + (1-\alpha) \hat{\lambda}_k(\tau) . \nonumber
\end{align}
This guarantees that all weighted gradients have the same variance over gradient components, i.e., that $\text{Var}[ \lambda_k  \nabla_{\bm{\theta}^{sh}} \mathcal{L}_k (\tau) ] \approx \max_{t=1\ldots K} \text{Var}[ \nabla_{\bm{\theta}^{sh}} \mathcal{L}_t (\tau) ]\: \forall k = 1,\ldots ,K $. Since the popular Adam optimizer is invariant to diagonal scaling of the gradients \cite{DBLP:journals/corr/KingmaB14}, the weights in Eq.~\ref{eq:varnorm} can in this case efficiently be computed as the inverse of averaged squared gradients of the task-specific objectives, which in turn is proportional to the Dirichlet energy of the objective, i.e., $ \mathbb{E}[ \left( \nabla_{\bm{\theta}^{sh}} \mathcal{L}_k \right)^2  ] \propto \int_{\Omega_{\bm{\theta}}} \vert \nabla_{\bm{\theta}^{sh}} \mathcal{L}_k \vert^2 \mathrm{d}\bm{\theta}$.
We therefore refer to this weighting strategy as \emph{inverse-Dirichlet weighting}. 

The weighting strategies in Eqs.~\ref{eq:max_avg} and \ref{eq:varnorm} are fundamentally different. Eq.~\ref{eq:max_avg} weights objectives in inverse proportion to \emph{certainty} as quantified by the average gradient magnitude. Eq.~\ref{eq:varnorm} uses weights that are inversely proportional to \emph{uncertainty} as quantified by the variance of the loss gradients. Equation~\ref{eq:varnorm} is also different from previous uncertainty-weighted approaches~\cite{kendall2018multi} because it measures the \emph{training uncertainty} stemming from noise in the loss gradients rather than the uncertainty from the model's \emph{observational noise}.

\subsection{Gradient-based multi-objective optimization} We compare weighting-based scalarization approaches with gradient-based multi-objective approaches using Karush-Kuhn-Tucker (KKT) local optimality conditions to find a descent direction that decreases all objectives. Specifically, we adapt the multiple gradient descent algorithm (MGDA) \cite{sener2018multi}, which leverages the KKT conditions 
\begin{equation}\label{eq:KKTconditions}
\exists \bm{\lambda} = \{\lambda_k \} \in \mathbb{R}_+^K \: \textrm{ s.t.} \sum_{k=1}^{K} \lambda_k \nabla_{\bm{\theta}^{sh}} \mathcal{L}_k (\bm{\theta}^{sh}, \bm{\theta}^k) = 0, \: \sum_{k=1}^{K} \lambda_k = 1
\end{equation}
to solve an MTL problem. The MGDA is guaranteed to converge to a Pareto-stationary point \cite{desideri2012multiple}. 
Given the gradients $\nabla_{\bm{\theta}^{sh}} \mathcal{L}_k$ of all objectives $k = 1,\ldots ,K$ with respect to the shared parameters, we find weights $\lambda_k$ that satisfy the above KKT conditions. The resulting $\bm{\lambda}$ leads to a descent direction that improves all  objectives~\cite{sener2018multi}, see also Suppl.~Fig.~\ref{fig:pareto_front}. We adapt the MGDA to PINNs as described in the Materials and Methods section. We refer to this adapted algorithm as \emph{pinn-MGDA}.

\section{Results}
As we posit above, PINNs are expected to fail for problems with dominant high frequencies, i.e, with high $\bm{k}_0$ or high derivatives $m$, as is typical for multi-scale PDE models. We therefore present results of numerical experiments for such cases and compare the approximation properties of PINNs with different weighting schemes. This showcases learning failure modes originating from vanishing task-specific gradients. 

We first consider Sobolev training of neural networks, which is a special case of PINN training, as shown above. Due to the linear nature of the Sobolev loss (Eq.~\ref{eq:Sobolev}), this test case allows us to compute $\epsilon$-optimal analytical weights (see Materials and Methods), providing a baseline for our comparison. Second, we consider a real-world example for a nonlinear PDE, using PINNs to model active turbulence.

In the first example, we specifically consider the Sobolev training problem with target function derivatives up to fourth order, i.e.~$K=4$, and total loss
\begin{equation}
 \sum_{i=1}^{N} \Bigg[ \big \vert  \bm{u}_{\bm{\theta}} (\bm{x}_i) - \bm{u}(\bm{x}_i) \big \vert^2  + \sum_{k=1}^{4}  \lambda_k \big \vert  \hat{\xi}_k  \mathcal{D}_{\textbf{x}}^k  \bm{u}_{\bm{\theta}} (\bm{x}_i) -\xi_k \mathcal{D}_{\textbf{x}}^k \bm{u}(\bm{x}_i) \big \vert^2 \Bigg ].
 \label{eq:exloss}
\end{equation}
The neural network is trained on the data $\bm{u}(\bm{x}_i)$ and their derivatives. To mimic inverse-modeling scenarios, in addition to producing an accurate estimate  $\bm{u}_{\bm{\theta}}(\bm{x}_i)$ of the function, the neural network is also tasked to infer unknown scalar prefactors $\bm{\hat{\xi}} = (\hat{\xi}_1,\hat{\xi}_2,\hat{\xi}_3,\hat{\xi}_4)$ with true values chosen as $\xi_{k} = 1$. 
This mimics scenarios in which unknown coefficients of the PDE model also needed to be inferred from data. For this loss, $\epsilon$-optimal weights can be analytically determined (see Materials and Methods) such that all objectives are minimized in an unbiased fashion~\cite{van2020optimally}. This provides the baseline against which we benchmark the different weighting schemes. 

\subsection{Inverse-Dirichlet weighting avoids vanishing task-specific gradients} 
We characterize the phenomenon of vanishing task-specific gradients, demonstrate its impact on the accuracy of a learned function approximation, and empirically validate Proposition \ref{prop}. For this, we consider a generic neural network with 5 layers and 64 neurons per layer, tasked with learning a 2D function that contains a wide spectrum of frequencies and unknown coefficients $\bm{\hat{\xi}} = (\hat{\xi}_1,\hat{\xi}_2,\hat{\xi}_3,\hat{\xi}_4)$ using Sobolev training with the loss from Eq.~\ref{eq:exloss}. The details of the test problem and the training setup are given in the Supplementary Material. 

\begin{figure}
\centering
\includegraphics[width=3.3in]{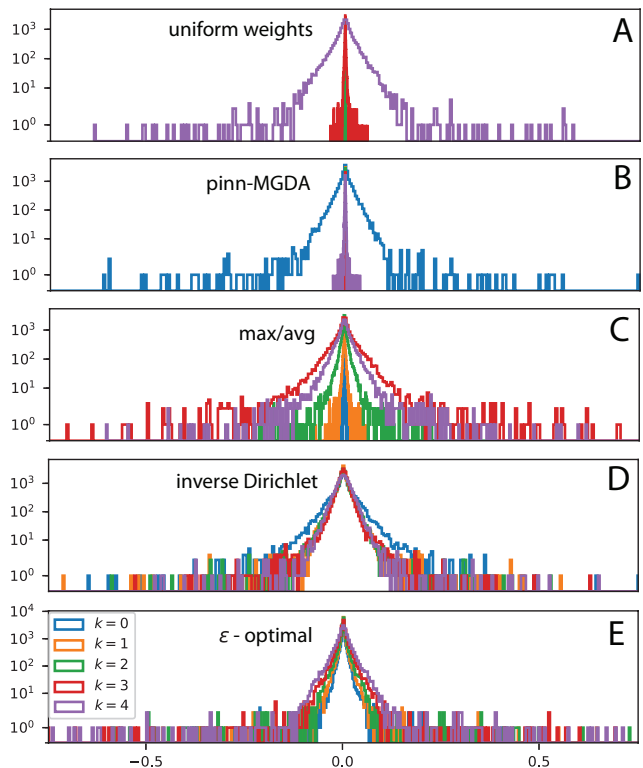}
\caption{\textbf{Gradient histograms illustrating vanishing task-specific gradients.}\footnotesize{ The gradient distributions of the different objectives ($k=0,\ldots ,4$, see main text) are shown in different colors (see inset legend) at training epoch 5000. The panels correspond to different $\lambda$ weighting schemes as named.
}}
\label{fig:sob_dist}
\end{figure}

We first confirm that in this test case vanishing task-specific gradients occur when using uniform weights, i.e., when setting $\lambda_k =1$ for all $k=0,\ldots , 4$. 
For this, we plot the gradient histogram of the five loss terms in Fig.~\ref{fig:sob_dist}A. The histograms clearly show that training is biased in this case  to mainly optimize the objective containing the highest derivative $k=4$. All other objectives are neglected, as predicted by Proposition \ref{prop}. This leads to a uniformly high approximation error on the test data (training vs.~test data split 50:50, see Supplementary Material), as shown in Fig.~\ref{fig:Sobolev_results}B, as well as a uniformly high error in the estimated coefficients $\bm{\xi}$ (Fig.~\ref{fig:Sobolev_results}C). When determining the weights using the Pareto-seeking pinn-MGDA, the accuracy considerably improves (Fig.~\ref{fig:Sobolev_results}B,C). However, the phenomenon of vanishing task-specific gradients is still present, as seen in Fig.~\ref{fig:sob_dist}B, this time favoring the objective for $k=0$ and neglecting the derivatives. Max/avg weighting as proposed in Ref.~\cite{wang2020understanding} and given in Eq.~\ref{eq:max_avg} improves the accuracy for higher derivatives (Fig.~\ref{fig:Sobolev_results}B), but still suffers from unbalanced gradient distributions (Fig.~\ref{fig:sob_dist}C), which leads to a uniformly high error in the estimated coefficients $\bm{\xi}$ (Fig.~\ref{fig:Sobolev_results}C). The inverse-Dirichlet weighting proposed here and detailed in Eq.~\ref{eq:varnorm} shows balanced gradient histograms (Fig.~\ref{fig:sob_dist}D) closest to those observed when using $\epsilon$-optimal analytical weights (Fig.~\ref{fig:sob_dist}E). This leads to orders of magnitude better accuracy in both the function approximation and the coefficient estimates, as shown in Fig.~\ref{fig:Sobolev_results}B,C, reaching the performance achieved when using $\epsilon$-optimal weights.

This is confirmed when looking at the power spectrum of the function approximation $\bm{u}_{\bm{\theta}}$ learned by the neural network in Fig.~\ref{fig:Sobolev_results}D. Inverse-Dirichlet weighting leads to results that are as good as those achieved by optimal weights in this test case, in particular to approximate high frequencies, as predicted by Proposition \ref{prop}.  A closer look at the power spectra learned by the different weighting strategies also reveals that the max/avg strategy fails to capture high frequencies, whereas the pinn-MDGA performs surprisingly well (Fig.~\ref{fig:Sobolev_results}D) despite its unbalanced gradients (Fig.~\ref{fig:sob_dist}B). This may be problem-specific, as the weight trajectories taken by pinn-MDGA during training (Suppl.~Fig.~\ref{fig:lambda_traj_sobolev}) are fundamentally different from the behavior of the other methods. This could be because the pinn-MGDA uses a fundamentally different strategy purely based on satisfying Pareto-stationarity conditions (Eq.~\ref{eq:KKTconditions}). As shown in Suppl.~Fig.~\ref{fig:wavespec_sobolev}, this leads to a rapid attenuation of the Fourier spectrum of the residual $\vert \tilde{\bm{r}}(\bm{k})\vert$ over the first few learning epochs in this test case, allowing pinn-MGDA to escape the stiffness defined in Proposition \ref{prop}.

In summary, we observe stark differences between different weighting strategies. Of the strategies compared, only the inverse-Dirichlet weighting proposed here is able to avoid vanishing task-specific gradients, performing on par with the $\epsilon$-optimal solution. The pinn-MGDA seems to be able to escape, rather than avoid, the stiffness caused by vanishing task-specific gradients. 
We also observe a clear correlation between having balanced effective gradient distributions and achieving good approximation and estimation accuracy. 

\begin{figure*}
\centering
\includegraphics[width=7.0in]{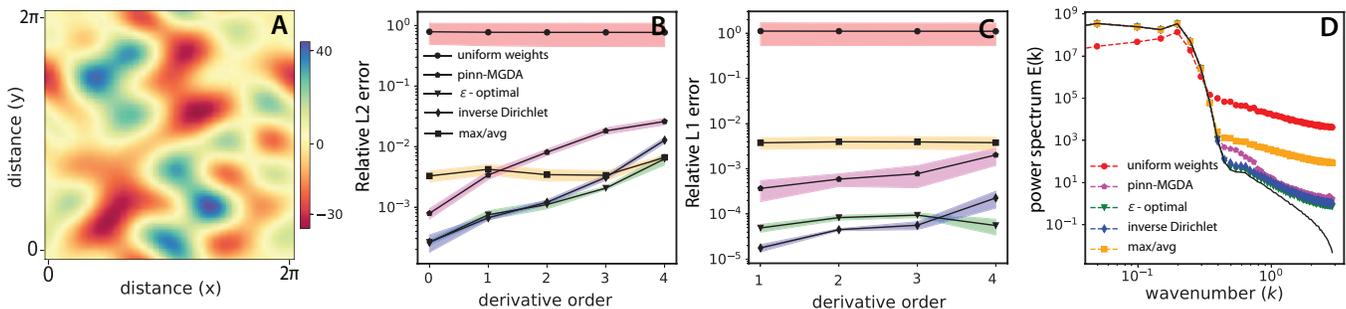}
\caption{\textbf{Approximation accuracy and convergence for Sobolev training using different weighting strategies.}\footnotesize{ A) One instance of the function $\bm{u}$ given by Eq.~$\ref{Eq:sinfunc}$
visualized on a $128 \times 128 $ regular Cartesian grid. B) The relative L2 error of the learned function approximation, i.e.~$\| \bm{u}_{\bm{\theta}}(\bm{x}_i) - \bm{u}(\bm{x}_i)\|_2/\|\bm{u}(\bm{x}_i) \|_2$, across all grid nodes $\bm{x}_i$ of the test data $\bm{u}(\bm{x}_i)$ for different objective weighting methods (symbols, see inset legend). C) The relative L1 error in the estimated model coefficients, i.e.~$\| \bm{\xi} - \bm{\hat{\xi}} \|_1/\|\bm{\xi}\|_1$ for the different objective weighting methods. D) The power spectrum $E(k)$ of the true function $\bm{u}$ (solid black line) compared with that of the learned estimators $\bm{u}_{\bm{\theta}}$ for the different weighting methods (colored lines with symbols). All results are shown after training for 20\,000 epochs using the Adam optimizer. The colored bands show to the standard deviation over 5 repetitions for different realizations (i.e., RNG seeds) of the random function $\bm{u}$ (see Suppl.~Eq.~\ref{Eq:sinfunc}). }}
\label{fig:Sobolev_results}
\end{figure*}

\subsection{Inverse-Dirichlet weighting enables multi-scale modeling using PINNs}

We investigate the use of PINNs in modeling multi-scale dynamics in space and time where vanishing task-specific gradients are a common problem. As a real-world test problem, we consider meso-scale active turbulence as it occurs in living fluids, such as bacterial suspensions \cite{wensink2012meso, dunkel2013fluid} and multi-cellular tissues \cite{ramaswamy2016activity}. Unlike inertial turbulence, active turbulence has energy injected at small length scales comparable to the size of the actively moving entities, like the bacteria or individual motor proteins. The spatiotemporal dynamics of active turbulence can be modeled using the generalized incompressible Navier-Stokes equation \cite{wensink2012meso,dunkel2013fluid}:
\begin{align}
        \frac{\partial \bm{u}}{\partial t} + \xi_0 (\bm{u} \cdot \nabla \bm{u}) &= - \nabla p + \xi_1 \nabla \vert \bm{u}\vert^2 - \alpha \bm{u} - \beta \vert \bm{u}\vert^2 \bm{u} + \notag \\
         &\qquad \Gamma_0 \Delta \bm{u} - \Gamma_2 \Delta^2 \bm{u} \label{eq:activeturb} \\
        \nabla \cdot \bm{u} &= 0 \notag,
\end{align}
where $\bm{u}(t,\bm{x})$ is the mean-field flow velocity and $p(t,\bm{x})$ the pressure field. The coefficient $\xi_0$ scales the contribution from advection and $\xi_1$ the active pressure. The $\alpha$ and $\beta$ terms correspond to a quartic Landau-type velocity potential. The higher-order terms with coefficients $\Gamma_0$ and $\Gamma_2$ model passive and active stresses arising from hydrodynamic and steric interactions, respectively~\cite{dunkel2013fluid}. Supplementary Figure~\ref{fig:activeturb_snapshot} illustrates the rich multi-scale dynamics observed when numerically solving this equation on a square for the parameter regime $\Gamma_0 < 0$ and $\Gamma_2 > 0$ \cite{wensink2012meso} (see Supplementary Material for details on the numerical methods used). An important characteristic of Eq.~\ref{eq:activeturb} is the presence of disparate length and time scales in addition to fourth-order derivatives and nonlinearities in $\bm{u}$, which jointly lead to considerable numerical stiffness \cite{cox2002exponential}. Given such large dominant wavenumbers $\bm{k}_0$, any data-driven approach, forward or inverse, is expected to encounter significant learning pathologies as per Proposition \ref{prop}.

We demonstrate these pathologies by applying PINNs to approximate the forward solution of Eq.~\ref{eq:activeturb} in 2D for the parameter regime explored in Ref.~\cite{wensink2012meso} in both square (Fig.~\ref{fig:activeturb_snapshot}) and annular (Fig.~\ref{fig:activeturbsolevol}E) domains. The details of simulation and training setup are given in the Supplementary Material. The loss function includes four terms as given in Eq.~\ref{eq:PINN}, corresponding to the PDE residual, initial and boundary conditions, and the divergence-freeness constraint.

Figure~\ref{fig:activeturbsolevol}A,B shows the average (over different initializations of the neural network weights) relative L2 errors of the vorticity field over training epochs when using different PINN weighting strategies. As predicted by Proposition~\ref{prop}, the uniformly weighted PINN fails completely both in the square (Fig.~\ref{fig:activeturbsolevol}A) and in the annular geometry (Fig.~\ref{fig:activeturbsolevol}B). In this case, optimization is entirely biased towards minimizing the PDE residual, neglecting the initial, boundary, and incompressibility constraints. This imbalance is also clearly visible in the loss gradient distributions shown in Suppl.~Fig.~\ref{fig:gradient_statistics}A. Comparing the approximation accuracy after 7000 training epochs confirms the findings from the previous section with inverse-Dirichlet weighting outperforming the other methods, followed by pinn-MGDA and max/avg weighting. This is made possible by assigning adequate weights for the initial and boundary conditions in relative to the equation residual (See  Suppl.~Fig.~\ref{fig:lamb_trajectory_turb}A,D).

In Fig.~\ref{fig:activeturbsolevol}C we compare the convergence of the learned approximation $\bm{u}_{\bm{\theta}}$ to the ground-truth numerical solution for different spatial resolutions $h$ of the discretization grid. The errors scale as $O(h^r)$ with convergence orders $r$ as identified by the dashed lines. While the pinn-MGDA and max/avg weighting achieve linear convergence ($r=1$), the inverse-Dirichlet weighted PINN converges with order $r=4$. The uniformly weighted PINN does not converge at all and is therefore not shown in the plot. 

Inverse-Dirichlet weighting also perfectly recovers the power spectrum $E(k)$ of the active turbulence velocity field, as shown in Fig.~\ref{fig:activeturbsolevol}D. This confirms that inverse-Dirichlet weighting enables PINNs to capture high frequencies in the PDE solution, corresponding to fine structures in the fields, by avoiding the stiffness problem from Proposition~\ref{prop}. Indeed, the PINNs trained with the other weighting methods fail to predict small structures in the flow field, as shown by the spatial distributions of the prediction errors in Fig.~\ref{fig:activeturbsolevol}F--I compared to the ground-truth numerical solution in Fig.~\ref{fig:activeturbsolevol}E. The evolution of the Fourier spectra $| \widetilde{r}(\bm{k}) |$ of the residuals of both velocity components $(u,v)=\bm{u}$ shown in Suppl.~Fig.~\ref{fig:wavespec_activeturb} confirm the rapid solution convergence achieved by inverse-Dirichlet weighted PINNs.

In the Supplementary Material, we provide results for two additional examples: approximation the solution of the 2D Poisson equation on a square with a large spectrum of frequencies, and learning the spatiotemporal dynamics of curvature-driven level-set flow developing small structures. In both supplementary multi-scale examples, we notice similar behavior with inverse-Dirichlet weighting outperforming the other approaches either in terms of accuracy or convergence rate.

In summary, our results suggest that avoiding vanishing task-specific gradients can enable the use of PINNs in multi-scale problems previously not amenable to neural-network modeling. Moreover, we observe that seemingly small changes in the weighting can lead to significant differences in the convergence order of the learned approximation.

\begin{figure*}[!t]
\centering
\includegraphics[width=7.0in]{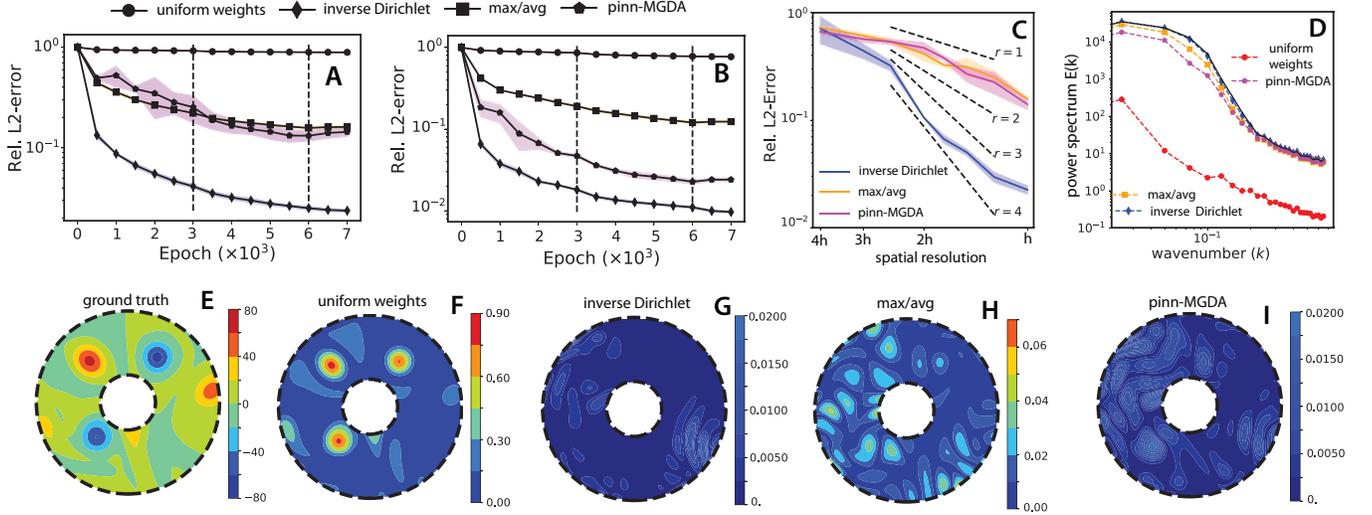}
\caption{\textbf{Forward modeling of active turbulence using PINNs.}\footnotesize{ A,B) Average relative L2 prediction errors for the vorticity field $\omega$ of the active turbulence PDE in Eq.~\ref{eq:activeturb} in square (A) and annular domains (B) using different weighting strategies for PINN training (symbols, see inset legend). C) Convergence of the different PINN predictions $\bm{u}_{\bm{\theta}}$ of the flow vorticity to the ground-truth numerical solution in a square domain for different grid resolutions. Dashed lines visualize integer convergence rates. The PINN solution is extracted at epoch $7000$. The colored bands show the standard deviations over three different initializations of the neural network weights.
D) Power spectrum $E(k)$ of the turbulent flow velocity field (solid black line) compared with the power spectra of the approximations learned by PINNs with different weighting (see inset legend) at epoch 500.
E) Visualization of the ground-truth vorticity ($\omega = \nabla\times \bm{u}$) field in an annular domain obtained by numerically solving Eq.~\ref{eq:activeturb} as detailed in the Supplementary Material.
F--I) Spatial distribution of the relative errors in the vorticity field when using PINNs with the weighting strategies named in the panel titles.}}
\label{fig:activeturbsolevol}
\end{figure*}

\subsection{Inverse-Dirichlet weighting protects against catastrophic forgetting}

Artificial neural networks tend to ``forget'' their parameterization when training for different tasks sequentially~\cite{kirkpatrick2017overcoming}, leading to a phenomenon referred to as \textit{catastrophic forgetting}. Sequential training is often used, e.g., when learning to de-noise data and estimate parameters at the same time, like combined image segmentation and denoising~\cite{buchholz2020denoiseg}, or when learning solutions of fluid mechanics models with additional constraints like divergence-freeness of the flow velocity field~\cite{jin2020nsfnets}.

In PINNs, catastrophic forgetting can occur, e.g., in the frequently practiced approach of first training on measurement data alone and adding the PDE model only at later epochs \cite{both2021deepmod}. Such sequential training can be motivated by the computational cost of evaluating the derivatives in the PDE using automatic differentiation, so it seems prudent to pre-condition the network for data fitting and introduce the residual later to regress for the missing network parameters.

\begin{figure*}
\centering
\includegraphics[width=6.8in]{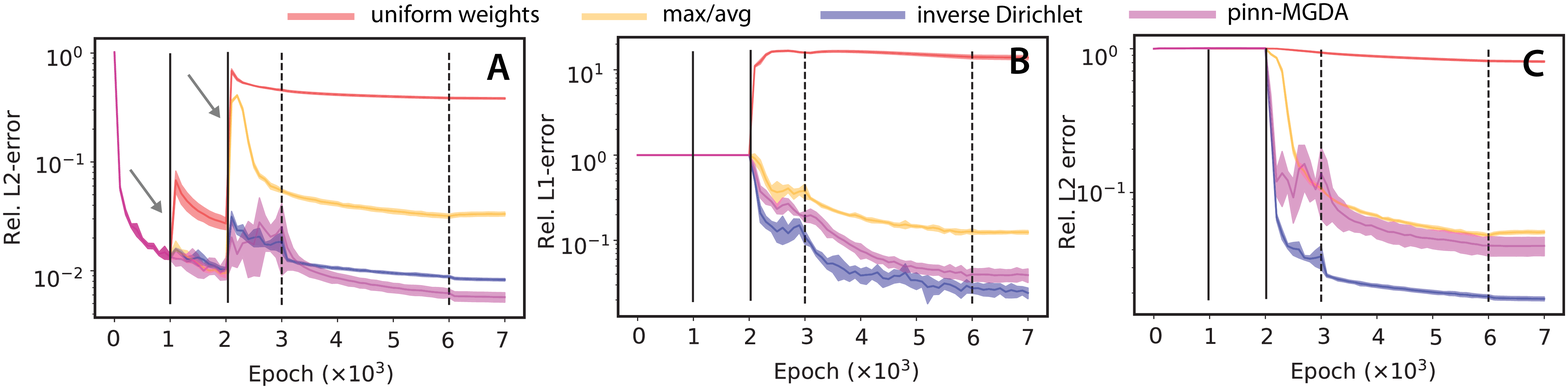}
\caption{\textbf{Catastrophic forgetting in sequential PINN training of active turbulence.}\footnotesize{ A): Test error in the velocity field $\bm{u}$ during the three-step sequential training described in the main text with different weighting strategies (colors). The solid vertical lines at epochs 1000 and  2000 mark the introduction of the divergence constraint and PDE residual, respectively. The dotted vertical lines mark epochs at which the learning rate $\eta(\tau)$ is adjusted. B) Average relative L1 errors in the inferred values for the coefficients $\xi_0$ ($\xi_1$ is absorbed into the effective pressure, see Supplementary Material), $\alpha$, $\beta$, $\Gamma_0$, and $\Gamma_2$ of Eq.~\ref{eq:activeturb} over training epochs. C) Average relative L2 error for inferring the latent effective pressure gradient $\nabla p^*$ across training epochs. 
}}
\label{fig:fisher}
\end{figure*}

Using the active turbulence test case described above, we investigate how the different weighting schemes compared here influence the phenomenon of catastrophic forgetting. For this, we consider the inverse problem of inferring unknown model parameters and latent fields like the effective pressure $p$ from flow velocity data $\bm{u}$. We therefore train a PINN in three sequential steps: (1) first fitting flow velocity data, (2) later including the additional objective for the incompressibility constraint $\nabla \cdot \bm{u} = 0$, (3) finally including also the objective for the PDE model residual. This creates a setup in which catastrophic forgetting occurs when using uniform PINN weights, as confirmed in Fig.~\ref{fig:fisher}A (arrows). As soon as additional objectives (divergence-freeness at epoch 1000 and equation residual at epoch 2000) are introduced, the PINN loses the parameterization learned on the previous objectives, leading to large and increasing test error for the field $\bm{u}$. 

Dynamic weighting strategies can protect PINNs against \textit{catastrophic forgetting}, by adequately adjusting the weights whenever an additional objective is introduced. This is confirmed in Fig.~\ref{fig:fisher}A with inverse-Dirichlet and pinn-MGDA only minimally affected by catastrophic forgetting. The max/avg weighting strategy~\cite{wang2020understanding}, however, does not protect the network quite as well. When looking at the test errors for the learned model coefficients (Fig.~\ref{fig:fisher}B) and the accuracy of estimating the latent pressure field $p(\bm{x},t)$ (Fig.~\ref{fig:fisher}C), for which no training data is given, the inverse-Dirichlet weighting proposed here outperforms the other approaches. 

In summary, inverse-Dirichlet weighting can protect a PINN against catastrophic forgetting, making sequential training a viable option. In our tests, it offered the best protection of all compared weighting schemes.

\section{Conclusion and Discussion}

Physics Informed Neural Networks (PINNs) have rapidly been adopted by the scientific community for diverse applications in forward and inverse modeling. It is straightforward to show that the training of a PINN amounts to a Multi-Task Learning (MTL) problem, which is sensitive to the choice of regularization weights. Trivial uniform weights are known to lead to failure of PINNs, empirically found to exacerbate for physical models with disparate scales or high derivatives. 

As we have shown here, this failure can be explained by analogy to numerical stiffness of Partial Differential Equations (PDEs) when understanding Sobolev training of neural networks~\cite{czarnecki2017sobolev} as a special case of PINN training. This connection enabled us to prove a proposition stating that PINN learning pathologies are caused by dominant high-frequency components or high derivatives in the physics prior. The asymptotic arguments in our proposition inspired a novel dynamic adaptation strategy for the regularization weights of a PINN during training, which we termed \emph{inverse-Dirichlet weighting}. 

We empirically compared inverse-Dirichlet weighting with the recently proposed max/avg weighting for PINNs~\cite{wang2020understanding}. Moreover, we looked at PINN training from the perspective of multi-objective optimization, which avoids {\it a-priori} choices of regularization weights. We therefore adapted to PINNs the state-of-the-art multiple gradient descent algorithm (MGDA) for training multi-objective artificial neural networks, which has previously been proven to converge to a Pareto-stationary solution~\cite{sener2018multi}. This provided another baseline to compare with. Finally, we compared against analytically derived $\epsilon$-optimal static weights in a simple linear test problem where those can be derived, providing an absolute baseline. In all comparisons, and across a wide range of problems from 2D Poisson to multi-scale active turbulence and curvature-driven level-set flow, inverse-Dirichlet weighting empirically performed best.

The inverse-Dirichlet weighting proposed here can be interpreted as an uncertainty weighting with training uncertainty quantified by the batch variance of the loss gradients. This is different from previous approaches~\cite{kendall2018multi} that used weights quantifying the model uncertainty, rather than the training uncertainty. The proposition proven here provides an explanation for why the loss gradient variance may be a good choice, and our results confirmed that weighting based on the Dirichlet energy of the loss objectives outperforms other weighting heuristics, as well as Pareto-front seeking algorithms like pinn-MGDA. The proposed inverse-Dirichlet weighting also performed best at protecting a PINN against catastrophic forgetting in sequential training settings where loss objectives are introduced one-by-one, making sequential training of PINNs a viable option in practice.

While we have focussed on PINNs here, our analysis equally applies to Sobolev training of other neural networks, and to other multi-objective losses,  as long as training is done using an explicit first-order stochastic gradient descent (SGD) optimizer, of which Adam~\cite{DBLP:journals/corr/KingmaB14} is most commonly used. It would also be interesting to see how the weighting strategies presented in this study generalize to other first-order optimizers \cite{schmidt2020descending}. 

Taken together, we have exploited a connection between PINNs and Sobolev training to explain a likely failure mode of PINNs, and we have proposed a simple yet effective strategy to avoid training failure. The proposed inverse-Dirichlet weighting strategy is easily included into any SDG-type optimizer at almost no additional computational cost (see Suppl.~Fig.~\ref{fig:timings}). It has the potential to improve the accuracy and convergence of PINNs by orders of magnitude, to enable sequential training with less risk of catastrophic forgetting, and to extend the application realm of PINNs to multi-scale problems that were previously not amenable to data-driven modeling. 

\begin{acknowledgments}
\noindent This work was supported by the German Research Foundation (DFG) -- EXC-2068, Cluster of Excellence ``Physics of Life'', and by the Center for Scalable Data Analytics and Artificial Intelligence (ScaDS.AI) Dresden/Leipzig, funded by the Federal Ministry of Education and Research (BMBF).
\end{acknowledgments}

\appendix

\bibliography{apssamp}

\counterwithin{figure}{section}

\section{Material and Methods}

\subsection{Determining analytical optimal weights}
For Sobolev loss functions based on elliptic PDEs, optimal weights $\lambda$ that minimize all objectives in an unbiased way can be determined by $\epsilon$-closeness analysis \cite{van2020optimally}.
\begin{defn} [$\epsilon$-closeness] \label{th:yarotsky} A candidate solution $\bm{u}$ is called $\epsilon$-close to the true solution $\hat{\bm{u}}$ if it satisfies 
\begin{equation}
\frac{\vert \mathcal{D}^{\bm{\beta}} \bm{u} (\mathbf{x}) -\mathcal{D}^{\bm{\beta}} \hat{\bm{u}} (\mathbf{x})\vert}{ \vert \mathcal{D}^{\bm{\beta}} \hat{\bm{u}}(\mathbf{x}) \vert}  \leq \epsilon 
\end{equation}
for any multi-index $\bm{\beta} \in \mathbb{N}$ and $\mathbf{x} \in \Omega \subset \mathbb{R}^d$, thus $\mathcal{D}^{\bm{\beta}}  \bm{u} (\mathbf{x}) = \frac{\partial^{\vert \bm{\beta} \vert} \bm{u}(\textbf{x})}{\partial x_1^{\beta_1}\partial x_2^{\beta_2}\ldots  \partial x_d^{\beta_d}}$ with $\vert \bm{\beta}\vert = \sum_{i=1}^{d} \beta_i$.
\end{defn}

\noindent For the loss function typically used in Sobolev training of neural networks, i.e.,
\begin{equation*}
    \mathcal{L}_k( \cdot ) = \int_{\Omega} \left( \mathcal{D}_{\mathbf{x}}^{k} \hat{\bm{u}} - \mathcal{D}_{\mathbf{x}}^{k} \bm{u}\right)^2 \mathrm{d} \bm{x}
\end{equation*}
we can use $\epsilon$-closeness analysis to show that
\begin{equation}
    \mathcal{L}_{k} \leq \epsilon^2 \int_{\Omega} \left( \vert \xi_k \mathcal{D}_{\bm{x}}^k (\bm{\hat{u}}) \vert \right)^2 \mathrm{d} \bm{x} \, .
\end{equation}
Using this inequality, we can bound the total loss as
\begin{equation}
\mathcal{L}(\bm{u}) \leq \epsilon^2 \sum_{k=1}^{K} \lambda_k \int_{\Omega} \left( \vert \xi_k \mathcal{D}_{\bm{x}}^k (\bm{\hat{u}}) \vert \right)^2 \mathrm{d} \bm{x} = \epsilon^2 \sum_{k=1}^{K} \lambda_k \mathcal{I}_k
\end{equation}
with $\mathcal{I}_k = \int_{\Omega} \left( \vert \xi_k \mathcal{D}_{\bm{x}}^k (\bm{\hat{u}}) \vert \right)^2\! \mathrm{d} \Omega $. This bound can be used to determine the weights $\{\lambda _k \} _{k=1}^K$ that result in the smallest $\epsilon$:
\begin{align}\label{eq:tightbound}
    \mathcal{L}_k(\bm{u}) & \leq \frac{1}{\lambda_k}  \mathcal{L}(\bm{u}) \leq \epsilon^2 \lambda_k \mathcal{I}_k \:\: \forall k = {1,\ldots ,K}, \nonumber\\
     \mathcal{L}(u) & \leq \epsilon^2 \min \{ \lambda_1 \mathcal{I}_1,\ldots ,\lambda_K \mathcal{I}_K \}.
\end{align}
The weights $\lambda_k, \forall k \in {1,...,K}$ that lead to the tightest upper bound on the total loss in Eq.~\ref{eq:tightbound} can be found by solving the maximization-minimization problem described in following corollary:
\begin{coro}
A maximization-minimization problem can be transformed into a piecewise linear convex optimization problem by introducing an auxiliary variable $z$, such that
\begin{align*}
    \textrm{maximize} \:\:z \qquad\quad & \quad  \\
   \textrm{subject to } \mathbf{1}^\top \bm{\lambda}^{*} = 1; \\  z \leq \lambda_k \mathcal{I}_k,  \quad k = {1,\ldots ,K} .
\end{align*}
The analytical solution to this problem exists and is given by $\lambda_k^* = \frac{\Pi_{j\neq k} \mathcal{I}_j}{\sum_{k=1}^{K}
 \Pi_{j\neq k} \mathcal{I}_j}, \quad k = 1,\ldots ,K$.
\end{coro}
For a Sobolev training problem with loss given by Eq.~\ref{eq:Sobolev}, weights that lead to $\epsilon$-optimal solutions are computed using $$\mathcal{I}_k = \int_{\Omega} \left( \vert \xi_k \mathcal{D}_{\bm{x}}^k (\bm{\hat{u}}) \vert \right)^2\! \mathrm{d} \bm{x}.$$\\

\subsection{PINN Multiple Gradient Descent Algorithm (pinn-MGDA)}
\noindent We adapt the MGDA to PINNs by formulating the conditions for Pareto stationary as stated in Eq.~\ref{eq:KKTconditions} as a quadratic program:
\begin{equation}\label{eq:QPproblem}
\underset{\bm{\lambda} \in \mathbb{R}^K}{\text{minimize }} \frac{1}{2} \bm{\lambda}^\top \bm{Q} \bm{\lambda}, \textrm{ s.t. } \lambda_k \geq 0 \: \forall k \in [K] \textrm{ and } \sum_{k=1}^K \lambda_k = 1,
\end{equation}
where $\bm{Q} = \bm{U}^\top\bm{U}$. The matrix $\bm{U} \in \mathbb{R}^{\vert \textrm{B} \vert \times K}$ is given by 
\begin{equation*}
\bm{U} = 
\begin{bmatrix}
          \vdots \:\: \qquad \qquad \vdots \qquad \qquad \:\: \vdots  \:\: \\
         \nabla_{\bm{\theta}^{sh}} \mathcal{L}_1 \:\: \nabla_{\bm{\theta}^{sh}} \mathcal{L}_2 \:\: \hdots  \:\:  \nabla_{\bm{\theta}^{sh}} \mathcal{L}_K \\
          \vdots \:\: \qquad \qquad \vdots \qquad \qquad \:\: \vdots  \:\: \\
\end{bmatrix}.
\end{equation*}
We solve the quadratic program in Eq.~\ref{eq:QPproblem} using the Frank-Wolfe algorithm.

\begin{figure}[ht!]
\centering
\includegraphics[width=2.5in]{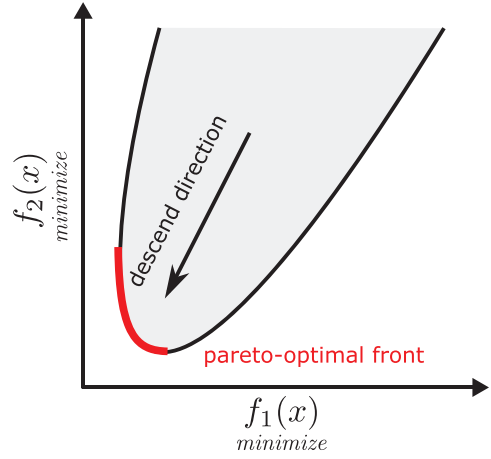}
\caption{\textbf{ Multiple gradient descent algorithm }\footnotesize{ The schematic of multi-objective optimization for minimizing two objectives $f_1(x)$ and $f_2(x)$. The red curve corresponds to the Pareto-optimal front. The MGDA algorithm searches for the descent direction (arrow) that minimizes all the objectives and is closer to the Pareto-optimal front. The grey region corresponds to the feasible objective space.}}
\label{fig:pareto_front}
\end{figure}

\clearpage
\onecolumngrid
\section{Sobolev Training}
\subsection*{Proof of Proposition 1}

\noindent Let us assume the Fourier spectrum of the signal is a delta function, i.e. $\widetilde{u}(k) = \Gamma_0 \delta (k - k_0)$, with $k_0$ being the dominant wavenumber. The residual at training time $(t)$ is given as  $r(\cdot,t) = u_\theta - u$, and its corresponding Fourier coefficients for the wavenumber $k$ is given by $\widetilde{r}(k,t) = \widetilde{u}_{\theta}(k,t) - \widetilde{u}(k)$. The learning rate of the residual for the dominant frequency is given as,
\begin{align}\label{eq:res_rate}
\Big \vert \frac{\partial \widetilde{r}(k_0,\tau)}{\partial \tau} \Big \vert &= \Big \vert \frac{\partial \widetilde{u}_{\theta}(k_0,\tau)}{\partial \theta}    \Big \vert \Big \vert \frac{\partial \theta}{\partial \tau} \Big \vert \nonumber \\
&= \Big \vert \eta \frac{\partial \widetilde{u}_{\theta}(k_0)}{\partial \theta}    \Big \vert  \Big \vert  \frac{\partial \mathcal{L}}{\partial \theta}  \Big \vert \nonumber \\
&= \Big \vert \eta \frac{\partial \widetilde{u}_{\theta}(k_0)}{\partial \theta}    \Big \vert  \Big \vert  \lambda_1 \frac{\partial \mathcal{L}_1}{\partial \theta}  + \lambda_2 \frac{\partial \mathcal{L}_2}{\partial \theta}  \Big \vert
\end{align}

\noindent From Parseval's theorem \cite{hardy1931note}, we can compute the magnitude of the gradients for both objectives $\mathcal{L}_1$ and $\mathcal{L}_2$ with respect to the network parameters $\theta$.
\begin{align}\label{eq:fun_part}
\frac{\partial \mathcal{L}_1}{\partial \theta} &= 2 \sum_{k=-N/2}^{N/2-1} \Big \vert \textbf{Re}\Big [ \widetilde{u}_{\theta} (k) - \widetilde{u}(k) \Big ] \frac{ \partial \widetilde{u_\theta} (k) }{\partial \theta} \Big \vert \nonumber \\
& \leq 2 \Big \vert  \Gamma_0 \frac{\partial \widetilde{u}_{\theta}(k_0)}{\partial \theta} \Big \vert + 2 \sum_{k=-N/2}^{N/2-1} \Big \vert \widetilde{u}_{\theta}(k)  \frac{\partial \widetilde{u}_{\theta} (k)}{\partial \theta } \Big \vert \nonumber \\ 
&\approx \mathcal{O}(1)  \Big \vert \frac{\partial \widetilde{r}(k_0)}{\partial \theta } \Big \vert 
\end{align}

\noindent The second step in the above calculation involves triangular inequality and the assumption that the Fourier spectrum of the signal is a delta function. In a similar fashion, we can repeat the same analysis for the Sobolev part of the objective function as follows,

\begin{align}\label{eq:sob_part}
\frac{\partial \mathcal{L}_2}{\partial \theta} &= 2 \sum_{k=-N/2}^{N/2-1} \Big \vert \textbf{Re} \Big [ (ik)^m  \big (\widetilde{u}_{\theta} (k) - \widetilde{u}(k) \big ) \Big ] \frac{ \partial (ik)^m  \widetilde{u}_{\theta}  (k) }{\partial \theta} \Big \vert \nonumber \\
& = 2 \sum_{k=-N/2}^{N/2-1} \Big \vert \textbf{Re} \Big [ (ik)^{2m}  \big (\widetilde{u}_{\theta} (k) - \widetilde{u}(k) \big) \Big ] \frac{ \partial \widetilde{u}_{\theta} (k) }{\partial \theta} \Big \vert \nonumber \\
& \leq 2 \Big \vert \Gamma_0 \textbf{Re} \big [  (ik_0)^{2m} \big ] \frac{\partial \widetilde{u}_{\theta}(k_0)}{\partial \theta} \Big \vert + 2 \sum_{k=-N/2}^{N/2-1} \Big \vert \textbf{Re} \big [(ik)^{2m} \big ]  \widetilde{u}_{\theta}(k) \frac{\partial \widetilde{u}_{\theta} (k)}{\partial \theta} \Big \vert \nonumber \\ 
& \approx  \mathcal{O} (k_0^{2m}) \Big \vert \frac{\partial \widetilde{r}(k_0)}{\partial \theta} \Big \vert 
\end{align}

\noindent By substituting equations (\ref{eq:fun_part}) and (\ref{eq:sob_part}) into equation (\ref{eq:res_rate}), we get,
\begin{align*}
\Big \vert \frac{\partial \widetilde{r}(k_0,\tau)}{\partial \tau} \Big \vert &=  \Big \vert \eta \frac{\partial \widetilde{u}_{\theta}(k_0)}{\partial \theta}    \Big \vert  \left(  \lambda_1 \mathcal{O}(1)  \Big \vert \frac{\partial \widetilde{r}(k_0)}{\partial \theta} \Big \vert + \lambda_2  \mathcal{O} (k_0^{2m}) \Big \vert \frac{\partial \widetilde{r}(k_0)}{\partial \theta} \Big \vert  \right) \\
&= \frac{\eta}{ \mathcal{O}(k_0)}   \Big[  \lambda_1 \mathcal{O}(1)   + \lambda_2  \mathcal{O} (k_0^{2m})   \Big] \Big \vert \frac{\partial \widetilde{r}(k_0)}{\partial \theta} \Big \vert
\end{align*}
We use the fact that the spectral rate of the network function residual is inversely proportional to the wavenumber, i.e. $\partial \widetilde{u}_\theta (k)/\partial \theta = \mathcal{O}\left(k^{-1}\right)$ \cite{rahaman2019spectral}.\\

\subsection*{Sobolev training set-up} \noindent We use a neural network $f_{\theta}: (x,y) \rightarrow (u)$ with 5 layers and 64 neurons per layer and $\sin(\cdot)$ activation function. The target function is a sinusoidal forcing term of the form,
\begin{equation}\label{Eq:sinfunc}
    u(x,y) = \sum_{i=1}^{M} A_x^i \cos(2\pi l_x^i x/L + \phi_x^i) A_y^i \sin(2\pi l_y^i y/L + \phi_y^i),
\end{equation}
where $M=20$ and $L=2\pi$ is the size of the domain. The parameters $A_x, A_y$ are drawn independently and uniformly from the interval $[-5,5]$, and similarly $\phi_x, \phi_y$ are drawn from the interval $[0, 2\pi]$. The parameters $l_x, l_y$ which set the local length scales are sampled uniformly from the set $\{1,2,3,4,5\}$. The target function $u(x,y)$ is given on a 2D spatial grid of resolution $128 \times 128$ covering the domain $[0, 2\pi] \times [0, 2\pi]$. The grid data is then divided into a $50/50$ train and test split which corresponds to $8192$ training points used. We train the neural network for 20000 epochs using the Adam optimizer \cite{kingma2014adam}. Each epoch is composed of 2 mini-batch updates of batch size $\vert \textrm{B} \vert =4096$. The initial leaning rate is chosen as $\eta=10^{-3}$ and decreased by a factor of $10$ after $10000$ and $15000$ epochs.\\

\begin{figure*}[ht!]
\centering
\includegraphics[width=5.0in]{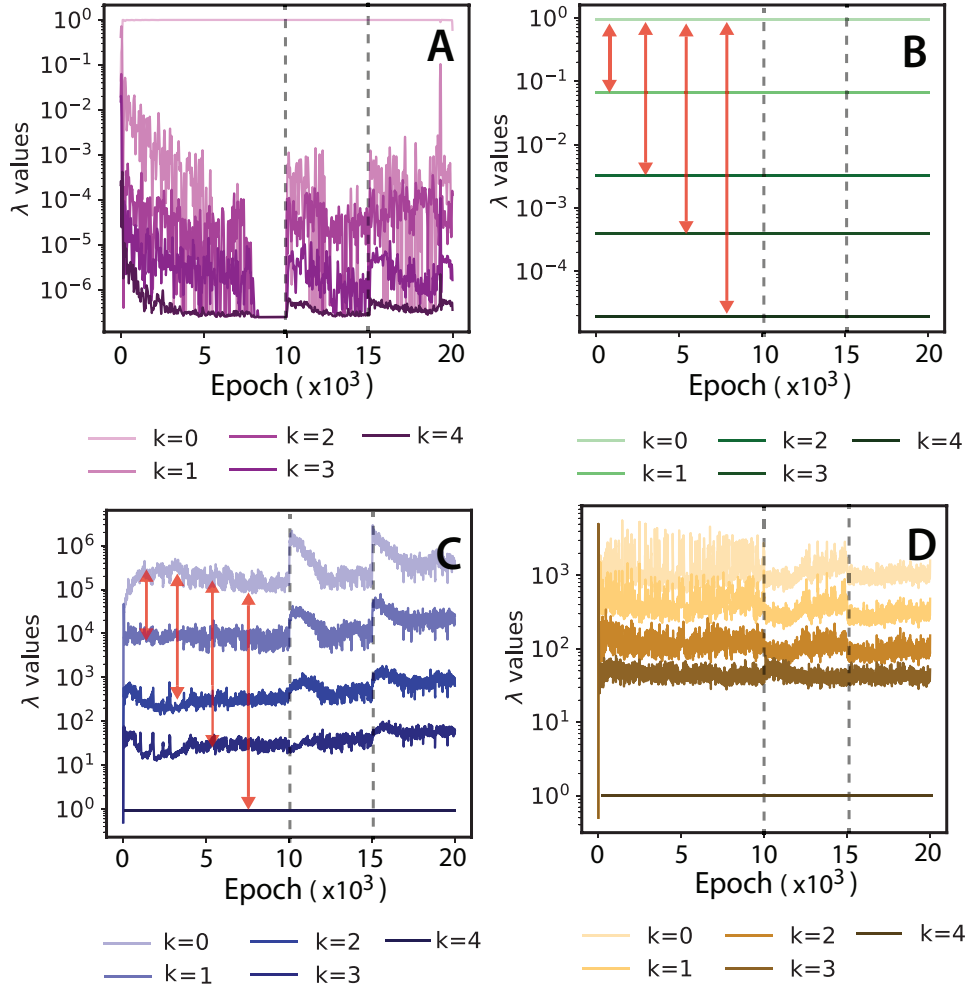}
\caption{\textbf{$\lambda$ trajectories for different weighing strategies }\footnotesize{ The dynamic weights $\lambda$ over the training epochs for a single instance of Sobolev training of the neural networks for different methods A) pinn-MGDA B) $\epsilon$-\textit{optimal} C) \textit{inverse-Dirichlet} D) \textit{max/avg}. Different trajectories in the same plots correspond to the weights $\lambda_k, k \in {1,..,K}$, associated with different objectives. On closer inspective we can notice that the relative weights or relative importance measure $(\lambda_i/\lambda_j) : \forall {i,j} \in {1,..,K}$ between objectives for \textit{inverse-Dirichlet} and the $\epsilon$-optimal strategy are approximately of the same order: $\lambda_0/\lambda_1 \approx 10, \lambda_0/\lambda_2 \approx 10^2 - 10^3, \lambda_0/\lambda_3 \approx 10^3 - 10^4 , \lambda_0/\lambda_4 \approx 10^4 - 10^5$ (shown in red double-headed arrows). }}
\label{fig:lambda_traj_sobolev}
\end{figure*}

\clearpage

\begin{figure*}[ht!]
\centering
\includegraphics[width=6.5in]{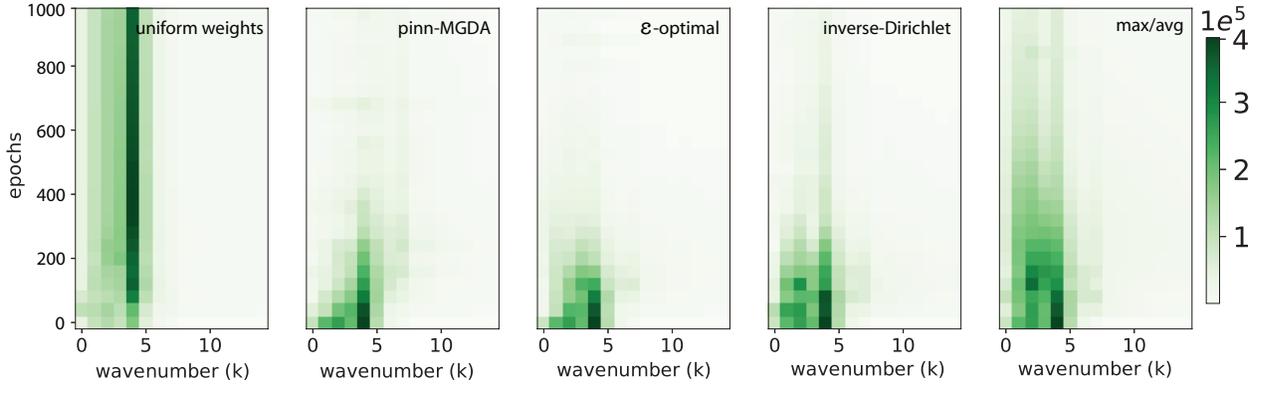}
\caption{\textbf{Evolution of the Fourier spectrum of the function residual $\widetilde{r}(\bm{k})$ }\footnotesize{ The Fourier spectrum of the residual tracked over the training epochs for different methods (shown in inset). The color intensity corresponds to the magnitude of the residual, i.e. $\vert \widetilde{r}(\bm{k}) \vert$.
}}
\label{fig:wavespec_sobolev}
\end{figure*}

\clearpage

\section{Forward and inverse modeling in active turbulence}
\noindent The ground truth solution of the 2D incompressible active fluid is computed in the vorticity formulation which is given as follows,
\begin{equation}\label{eq:}
\frac{\partial \omega}{\partial t} + \xi_0 \bm{u} \cdot \nabla \omega = -(\alpha + \beta \vert \bm{u}\vert^2) \omega + \beta \nabla \vert \bm{u} \vert^2 \times \bm{u} + \Gamma_0 \Delta \omega - \Gamma_2 \Delta^2 \omega.
\end{equation}


\noindent The simulation is conducted with a pseudo-spectral solver, where the derivatives are computed in the Fourier space and the time integration is done using the Integration factor method \cite{cox2002exponential, kassam2005fourth}. The equations are solved in the domain $[0, 2\pi] \times [0, 2\pi]$
with a spatial resolution of $256 \times 256$ and a time-step $dt=0.0001$. The incompressiblity is enforced by projecting the intermediate velocities on to the divergence-free manifold. 
The simulations in the square domain are performed with periodic boundary conditions, where the values of the parameters are chosen as follows: $\xi_0 = 3.5, \alpha = -1, \beta = 0.5, \Gamma_0 = -0.045, \Gamma_2 = \vert \Gamma \vert^3$ \cite{wensink2012meso}.

\subsection*{Forward problem set-up} We study the forward solution of the active turbulence problem using PINNs in the $(\omega-\psi)$ formulation. We employ a neural network $f_{\theta}: (x,y,t) \rightarrow (u,v)$ with 7 layers and 100 neurons per layer and $\sin$-activation functions. The vorticity $\omega$ is directly computed from the network outputs using automatic differentiation. For the solution in the square domain, we select a sub-domain $[\frac{\pi}{4},\frac{3\pi}{4}]\times [\frac{\pi}{4},\frac{3\pi}{4}]$ with 50 time steps  ($dt =0.01$) accounting to a total time of $0.5$ units in simulation time. We choose $N_{I} = 4096$ points for the initial conditions, $N_B = 12800$ boundary points and a total of $N_\text{int} = 192200$ residual points in the interior of the square domain. The network is trained for 7000 epochs using the Adam optimizer with each epoch consisting of $49$ mini-batch updates with a batch size $\vert\textrm{B}\vert = 4000$. The velocities are provided at the boundaries of the training domain.\\

\noindent For the forward simulation in the annular geometry, the domain is chosen with the inner radius of $r_\text{inner} \approx 0.245$ units and the outer radius $r_\text{outer}\approx 0.85$ units. We choose $N_{I} = 3536$ points for the initial conditions, $N_B = 14400$ boundary points and a total of $N_\text{int} = 162400$  residual points in the interior of the domain. The network training is again performed over 7000 epochs using Adam with each epoch consisting of $41$ mini-batch updates with a batch size $\vert\textrm{B}\vert = 4000$. Dirichlet boundary conditions are enforced at the inner and outer rims of the annular geometry by providing the velocities. The network is optimized subject to the loss function

\begin{equation*}
\mathcal{L}(\bm{u}, \theta) = \lambda_1 \mathcal{L}_\text{boundary} + \lambda_2 \mathcal{L}_\text{res} + \lambda_3  \mathcal{L}_\text{div} + \lambda_4 \mathcal{L}_\text{init}.
\end{equation*}
Here, $\lambda_1, \lambda_2, \lambda_3, \lambda_4$ correspond to the weights of the objectives for boundary condition, equation residual, divergence, and initial condition, respectively.
\begin{figure}[ht!]
\centering
\includegraphics[width=4.0in]{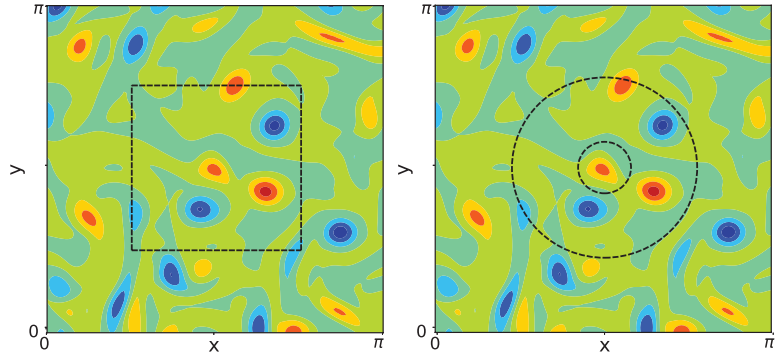}
\caption{\textbf{ Training domain for active turbulence problem }\footnotesize{ Left) Training domain (inside the dotted square) used for forward solution in the square domain. Right) Training domain (interior of the dotted annular outline) used for forward solution in the annular geometry.
}}
\label{fig:activeturb_snapshot}
\end{figure}
\subsection*{Inverse problem set-up} 
\begin{figure*}[ht!]
\centering
\includegraphics[width=7.0in]{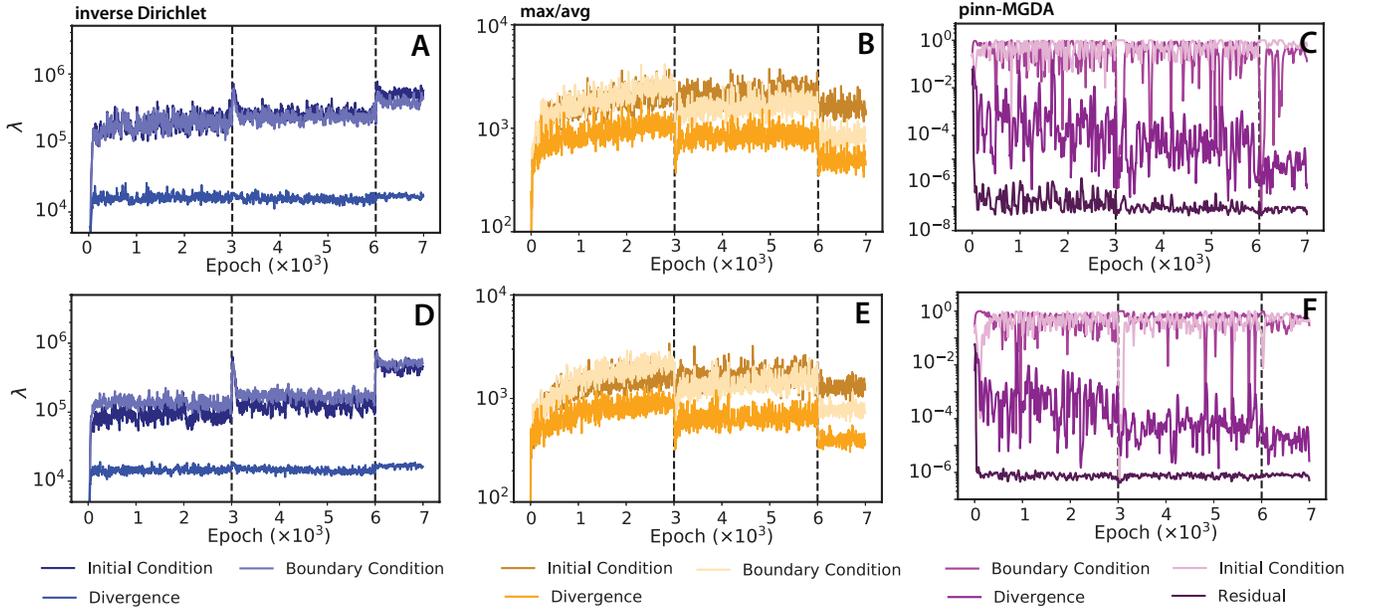}
\caption{\textbf{$\lambda$ trajectories for different weighting strategies for the active turbulence problem.}\footnotesize{ A, B, C) correspond to the $\lambda$ trajectories for the forward solution of the active turbulence problem in the square domain (See left Fig.~\ref{fig:activeturb_snapshot}). D,E,F) correspond to the $\lambda$ trajectories for the forward solution in the annular geometry (See right Fig.~\ref{fig:activeturb_snapshot}). The columns correspond to the different weighting strategies as named on the top left corner of each column.
}}
\label{fig:lamb_trajectory_turb}
\end{figure*}

\noindent For studying the inverse problem we resort to the primitive formulation $(u,v,p^{*})$  of the active turbulence model, with $p^{*}=-\nabla p+\xi_1\nabla |\bm{u}|^2$ as the effective pressure. We used a neural network $f_\theta:(x,y,t)\rightarrow (u,v,p^*)$ with 5 layers and 100 neurons per layer and $\sin$-activation functions. Given the measurement data of flow velocities $\{u_i, v_i \}_{i=1}^{N}$ at discrete points, we infer the effective pressure $p^{*}$ along with the parameters $\Sigma = \{ \xi_0, \alpha, \beta, \Gamma_0, \Gamma_2 \}$. The total of $204800$ points, sampled from the same domain as for the forward solution in the square domain, are utilized with a $70/30$ train and test split, which corresponds to $143360$ training points used. The network is trained for a total of $7000$ epochs using the Adam optimizer with each epoch consisting of $35$ mini-batch updates with a batch-size $\vert\textrm{B}\vert=4096$. We optimize the model subject to the loss function
\begin{equation*}
\mathcal{L}(\bm{u}, \theta) = \lambda_1 \mathcal{L}_\text{fit} + \lambda_2 \mathcal{L}_\text{res} + \lambda_3  \mathcal{L}_\text{div},
\end{equation*}
with $\lambda_1, \lambda_2, \lambda_3$ corresponding to the data-fidelity, equation residual and the objective that penalises non-zero divergence in velocities, respectively. The training is done in a three-step approach, with the first $1000$ epochs used for fitting the measurement data (pre-training, $\lambda_1 = 1, \lambda_2 = \lambda_3 = 0$), the next $1000$ epochs training with an additional constraint on the divergence ($\lambda_1  \neq 0, \lambda_2 \neq 0 , \lambda_3 = 0$), and finally followed by the introduction of the equation residual ($\lambda_1  \neq 0, \lambda_2 \neq 0 , \lambda_3 \neq 0$).  At every checkpoint introducing a new task, we reset all $\lambda_k$ for the computation of $\hat{\lambda}_k$ (see Eq. 7 main text) of the max/avg variant to $\lambda_k=1$. For both, forward and inverse problems, we choose an initial learning rate $\eta=10^{-3}$ and decrease it by a factor of $10$ after $3000$ and $6000$ epochs (See Fig.~\ref{fig:lamb_trajectory_turb}).

\clearpage

\begin{figure*}[ht!]
\centering
\includegraphics[width=6.5in]{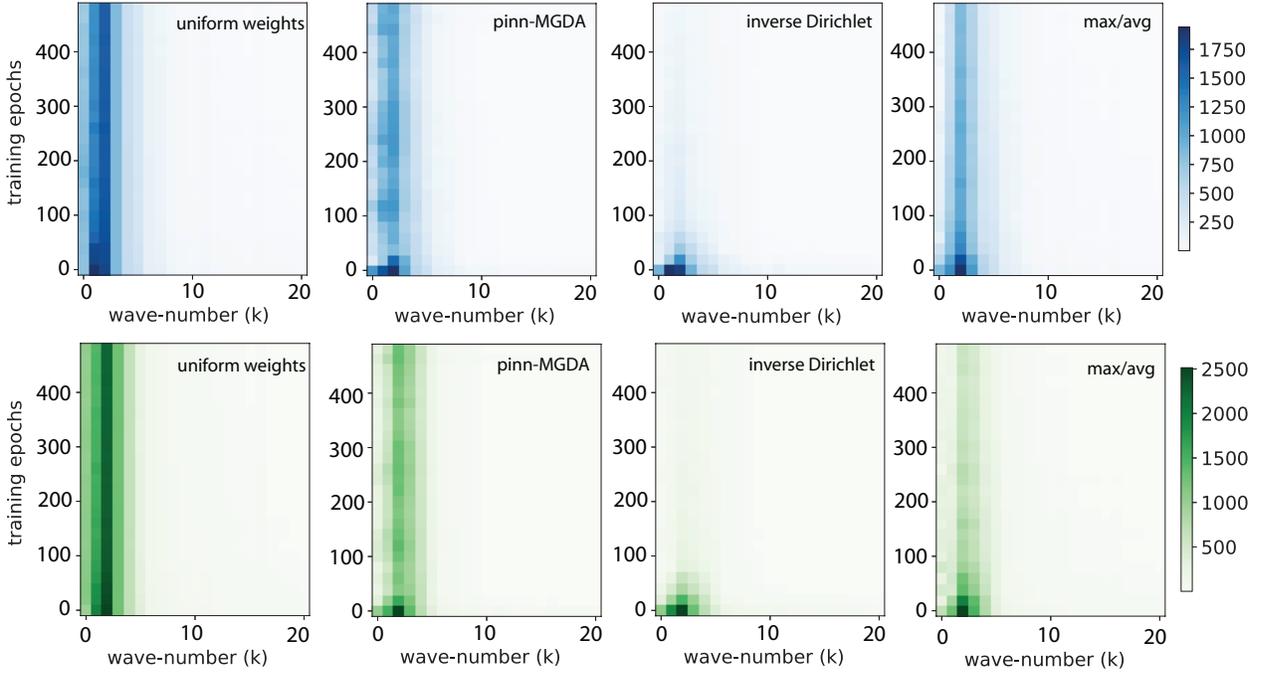}
\caption{\textbf{Evolution of Fourier spectrum of the function residual $\widetilde{r}(\bm{k})$. }\footnotesize{ Top row) The Fourier spectrum of the residual for the horizontal velocity component $u$. Bottom row) The Fourier spectrum of the residual for the vertical velocity component $v$. The columns correspond to different methods (shown in the inset). The color intensity corresponds to the magnitude of the residual, i.e. $\vert \widetilde{r}(\bm{k}) \vert$.
}}
\label{fig:wavespec_activeturb}
\end{figure*}

\begin{figure*}[ht!]
\centering
\includegraphics[width=4.0in]{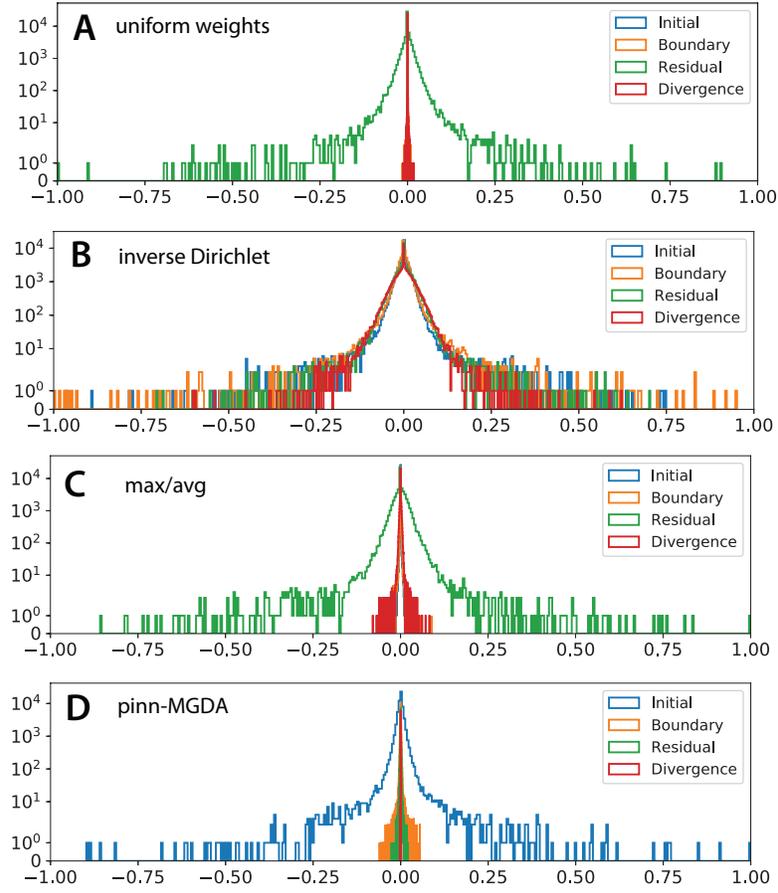}
\caption{\textbf{  Gradient histograms illustrating vanishing taks-specific gradients for the active turbulence problem.  }\footnotesize{The gradient distributions of the different objectives (insets on the right) at training epoch $5000$ for the forward problem in a squared domain. The panels correspond to different $\lambda$ weighting schemes as named.}}
\label{fig:gradient_statistics}
\end{figure*}


\begin{figure*}[ht!]
\centering
\includegraphics[width=6.5in]{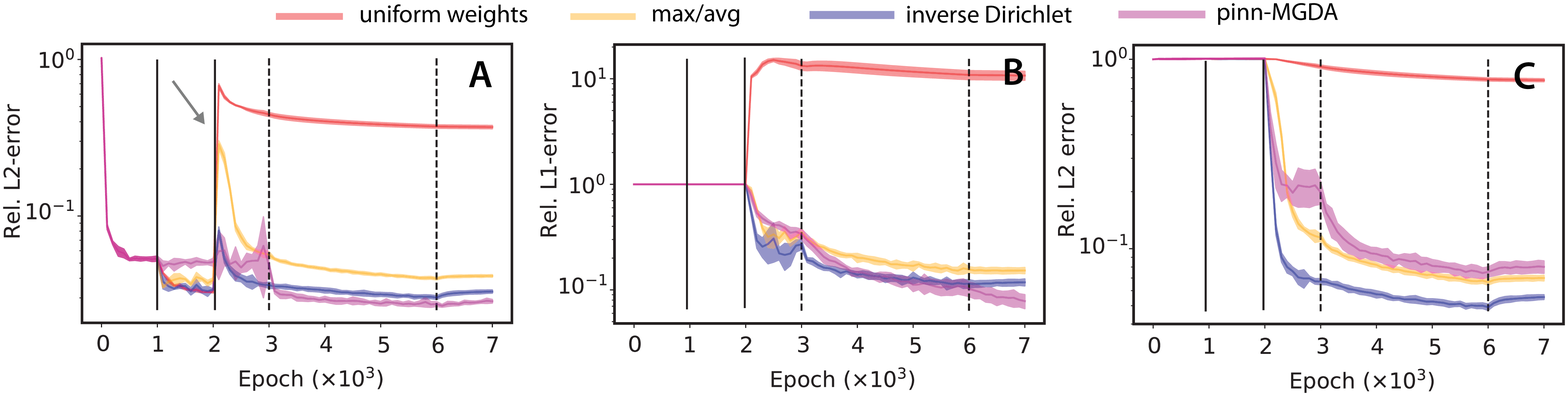}
\caption{\textbf{Catastrophic forgetting in sequential PINN training of active turbulence from noisy measurement data.}\footnotesize{ A): Test error in the velocity field $\bm{u}$ during the three-step sequential training described in the main text with different weighting strategies (colors). The solid vertical lines at epochs 1000 and  2000 mark the introduction of the divergence constraint and PDE residual, respectively. The dotted vertical lines mark epochs at which the learning rate $\eta(\tau)$ is adjusted. B) Average relative L1 errors in the inferred values for the coefficients $\xi_0$ ($\xi_1$ is absorbed into the effective pressure, see Supplementary Material), $\alpha$, $\beta$, $\Gamma_0$, and $\Gamma_2$ of Eq. 11 over training epochs. C) Average relative L2 error for inferring the latent pressure gradient $\nabla p$ across training epochs. The simulated data $\bm{u}$ is corrupted with additive Gaussian noise $\hat{\bm{u}} = \bm{u} + 0.25  \bm{\epsilon}$, where $\bm{\epsilon}\sim\mathcal{G}(0,\psi^2)$ is a vector of elementwise independent and identically distributed Gaussian random numbers with mean zero and empirical variance $\psi^2 = \textrm{Var}\{u_1,...,u_N \}$.
}}
\label{fig:losses_catastrophic}
\end{figure*}

\begin{figure*}[ht!]
\centering
\includegraphics[width=7in]{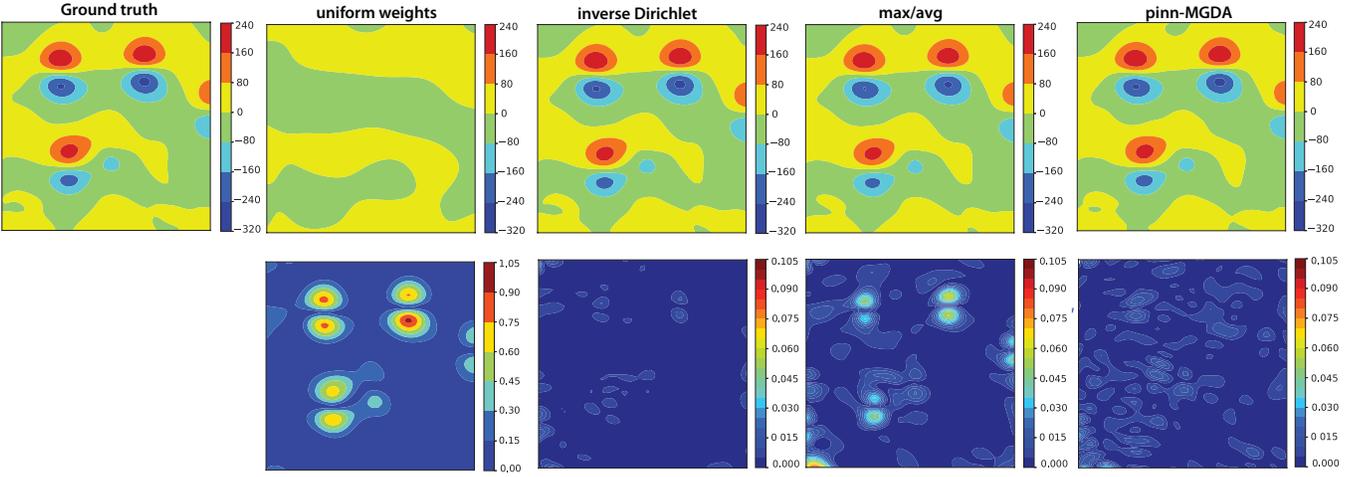}
\caption{\textbf{Inferred effective pressure.}\footnotesize{ The top row corresponds to the prediction of horizontal component of the effective pressure gradient $\partial_x p^*$ given the velocity field $\bm{u}$. The bottom row shows the corresponding point-wise relative error for the prediction. Different columns correspond to different weighting strategies as per the titles.
}}
\label{fig:pressure_snapshot}
\end{figure*}

\begin{figure*}[ht!]
\centering
\includegraphics[width=7in]{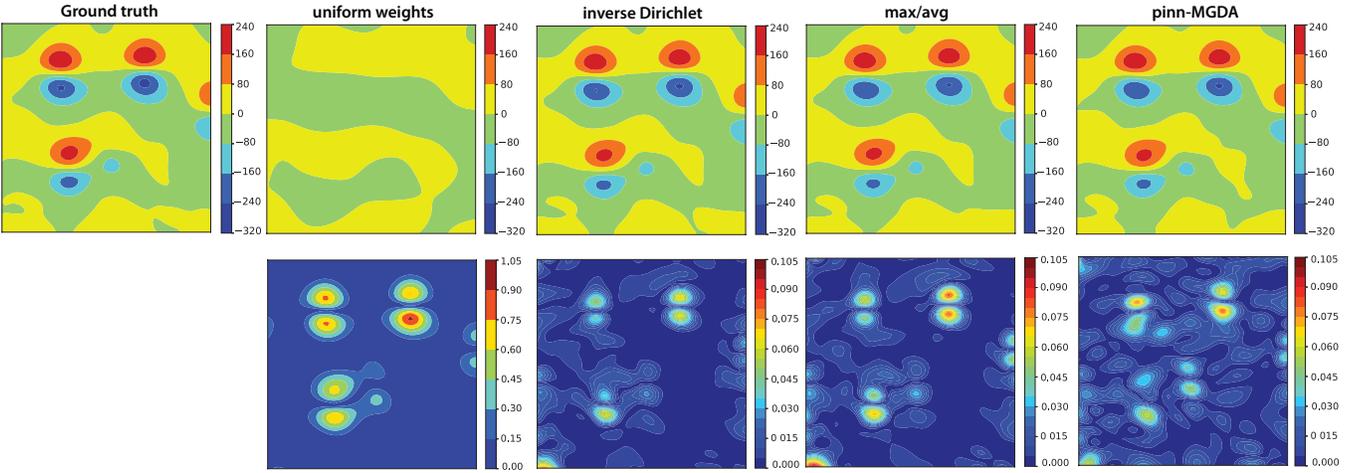}
\caption{\textbf{Inferred effective pressure from noisy measurement data.}\footnotesize{ The top row corresponds to the prediction of horizontal component of the effective pressure gradient $\partial_x p^*$ given the noise corrupted velocity field  $\hat{\bm{u}} = \bm{u} + 0.25  \bm{\epsilon}$, where $\bm{\epsilon}\sim\mathcal{G}(0,\psi^2)$ is a vector of elementwise independent and identically distributed Gaussian random numbers with mean zero and empirical variance $\psi^2 = \textrm{Var}\{u_1,...,u_N \}$. The bottom row shows the corresponding point-wise relative error for the prediction. Different columns correspond to different weighting strategies (titles).
}}
\label{fig:pressure_snapshot_noise}
\end{figure*}

\clearpage

\begin{figure*}[ht!]
\centering
\includegraphics[width=6.0in]{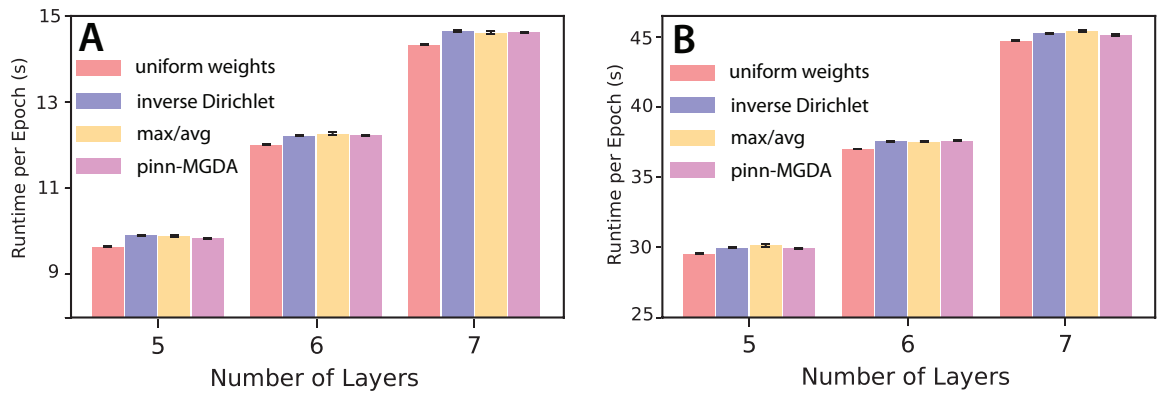}

\caption{\textbf{ Computational aspects of dynamic weighing for the active turbulence problem }\footnotesize{ Left) Training times per epoch for the inverse problem in the primitive formulation with $35$ mini-batch updates per epoch of batch size $\vert\textrm{B}\vert =4096$. Right) Training times per epoch for the forward problem in the vorticity-streamline formulation with $49$ mini-batch updates per epoch of batch size $\vert\textrm{B}\vert =4000$. The extra computational cost of the forward problem comes from using more training points and from the additional derivative that needs to be computed for evaluating the vorticity. The profiling was done on a NVIDIA GTX 1080 graphics card equipped with 8GB VRAM.
}}
\label{fig:timings}
\end{figure*}

\clearpage

\onecolumngrid
\section{2D Poisson problem}
\begin{figure*}[ht!]
\centering
\includegraphics[width=5.0in]{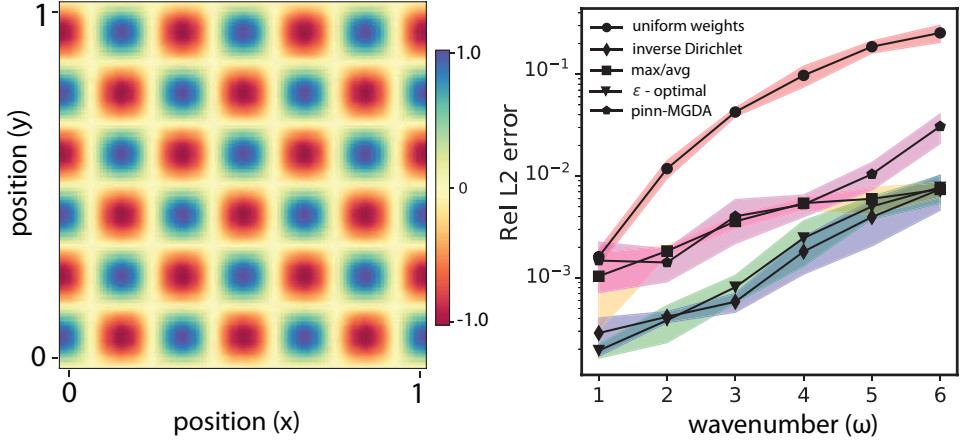}
\caption{\textbf{Comparison between methods for solution of Poisson Equation }\footnotesize{ Left: Solution of the Poisson equation (Eq.~\ref{Eq:poisson}) for the wave-number $\omega=6$. Comparison between relative L2 error for different weighing strategies for increasing wave-numbers $(\omega)$.
}}
\label{fig:Poisson}
\end{figure*}

\noindent We consider the problem of solving the 2D Poisson equation given as,
\begin{equation}\label{Eq:poisson}
    \Delta u = -2\omega^2 \cos(\omega x) \sin(\omega y),
\end{equation}
with the parameter $\omega$ controlling the frequency of the oscillatory solution $u(x,y)$. The analytical solution $u(x,y) = \cos(\omega x)\sin(\omega y)$ is computed on the domain $[0,1]\times[0,1]$ with a grid resolution of $100\times 100$. The boundary condition is given as $\mathcal{B}(x,y)=\cos{(\omega x)}\sin{(\omega y)}$, $x,y\in\partial\Omega$. We set up the PINN with two objective functions one for handling boundary conditions and the other to compute the residual with the loss
\begin{equation}
    \mathcal{L}(u, \theta) = \lambda_1 \mathcal{L}_\text{res} + \lambda_2 \mathcal{L}_\text{boundary}
\end{equation}
\noindent We set $N_\mathcal{B} = 400$ sampled uniformly on the four boundaries, and randomly sample $N_\text{int} = 2500$ points from the interior of the domain. We choose a network $f_\theta: (x,y) \rightarrow (u)$ with $5$ layers and $50$ nodes per layer and the $\tanh$ activation function is used. The neural network is trained for $30000$ epochs using the Adam optimizer. We choose the initial learning rate $\eta=10^{-3}$ and decrease it by a factor of $10$ after $10000$, $20000$ and $30000$ epochs. In Fig.~\ref{fig:Poisson} we show the relative L2 error for different weighting strategies across changing wavenumber $\omega$. We notice a very similar trend as observed in the previous problems, with \textit{inverse Dirichlet} scheme performing in par with the analytical $\epsilon$ optimal weighting strategy, followed by max/avg and pinn-MGDA.  \\


\clearpage
\onecolumngrid
\section{Curvature-driven level-set}
\noindent The level-set method \cite{osher2006level} is a popular method used for implicit representation of surfaces. In contrast to explicit methods for interface capturing, level-set methods offer a natural means to handle complex changes in the interface topology. For this reason, level-set methods have found extensive applications in computer vision \cite{malladi1995image}, multiphase and computational fluid dynamics \cite{zhao1996variational}, and for solving PDEs on surfaces \cite{bergdorf2010lagrangian}. For this example, we consider the curvature driven flow problem that evolves the topology of a complex interface based on its local curvature, and can be tailored for image processing tasks like image smoothing and enhancement \cite{malladi1995image, alvarez1993axioms}. Specifically, we look at the problem of a wound spiral which relaxes under its own curvature \cite{osher2006level}. The parametrization of the initial spiral is given by $\theta = 2 \pi D \sqrt{s}$ with $s = (k+a)/(np+a)$.  The point locations of the zero contour can be computed as,
\begin{align*}
x_s &= x_c + m \left( D/(1+D) \sqrt{s} \cos(\theta) \right) \\
y_s &= y_c + m \left( D/(1+D) \sqrt{s} \sin(\theta) \right)
\end{align*}

\noindent The zero level set function can then be computed using the distance function $d(\bm{x}) = \Vert \bm{x}- \bm{x}_p\Vert$ as $\phi(\bm{x},t=0) = d(\bm{x}) - w$ with $w$ being the width of the spirals. The distance function $d(\bm{x})$ measures the shortest distance between two points $\bm{x}$ and $\bm{x}_p$. Given the level set function $\phi(\bm{x})$, we can compute the 2D curvature $\kappa$ as,
\begin{equation}
\kappa = \frac{\phi_{xx} \phi_y \phi_y - 2\phi_{xy} \phi_x \phi_y + \phi_{yy} \phi_x \phi_x}{ \left( \phi_x^2 + \phi_y^2 \right)^{3/2}}
\end{equation}

\noindent The evolution of the level set function driven by the local curvature $\kappa$ and is computed using the PDE,
\begin{equation*}
\frac{\partial \phi}{\partial t} = \kappa \vert \nabla \phi \vert
\end{equation*}

\noindent The simulation was conducted on an $128 \times 128$ resolution spatial grid covering the domain $[-2,2]\times[-2,2]$ using the finite-difference method with necessary reinitialization to preserve the sign distance property of the level-set function $\phi$. The spiral parameters are chosen to be $np=400, D=2.5, a = 3, m = 2$ \cite{alame2020variational}. Time integration is performed using a second-order Runge-Kutta scheme with a step size of  $dt=0.001$.

\subsection*{Training set-up} The training data is generated from uniformly sampling in the domain $[-1.5,1.5] \times [-1.5,1.5]$ with a resolution of $100 \times 100$ in space and $80$ time snapshots separated with time-step $dt=0.001$. This lead to $N_I = 10000, N_B = 32000$ and $N_\text{int} = 768000$ points in the interior of the space and time domain. The network is trained for $2500$ epochs using the Adam optimizer with each epoch consisting of $187$ mini-batch updates with batch size $\vert\textrm{B}\vert =4096$. The neural network $f_\theta: (x,y,t) \rightarrow (\phi)$ with $5$ layers and $64$ nodes is chosen for this problem. We found that using the ELU activation function resulted in the best performance. The total loss is given as,
\begin{equation}
\mathcal{L}(\bm{u}) = \lambda_1 \mathcal{L}_\text{boundary} + \lambda_2 \mathcal{L}_\text{res} + \lambda_3 \mathcal{L}_\text{init}
\end{equation}

\noindent We choose the initial learning rate as $\eta=10^{-3}$ and decrease it after $1500$ and $2000$ epochs by a factor of $10$.

\begin{figure*}[ht!]
\centering
\includegraphics[width=5.0in]{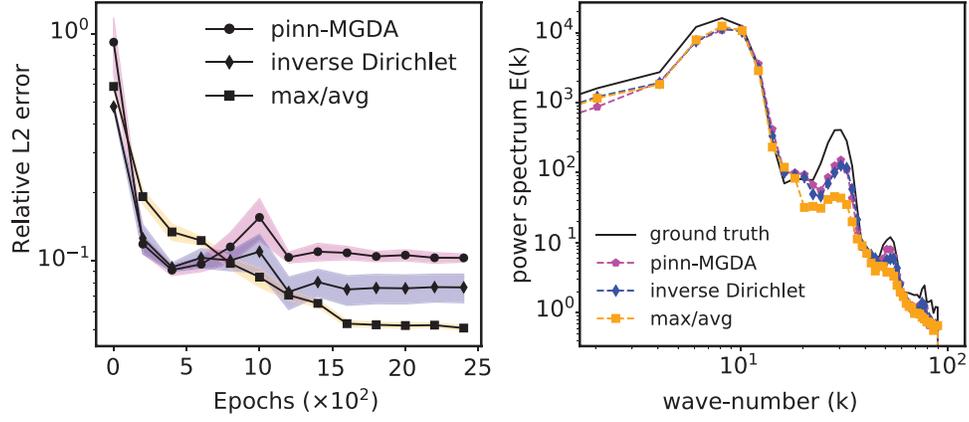}
\caption{\textbf{Curvature driven level-set problem}\footnotesize{ A) Relative L2 error for for different methods for the curvature driven level-set problem. B) The Power spectrum of the PINN level-set function approximation after 2000 epochs compared to the Power spectrum of the ground-truth (shown as solid black line).
}}
\label{fig:levelset_results}
\end{figure*}

\begin{figure*}[ht!]
\centering
\includegraphics[width=7.0in]{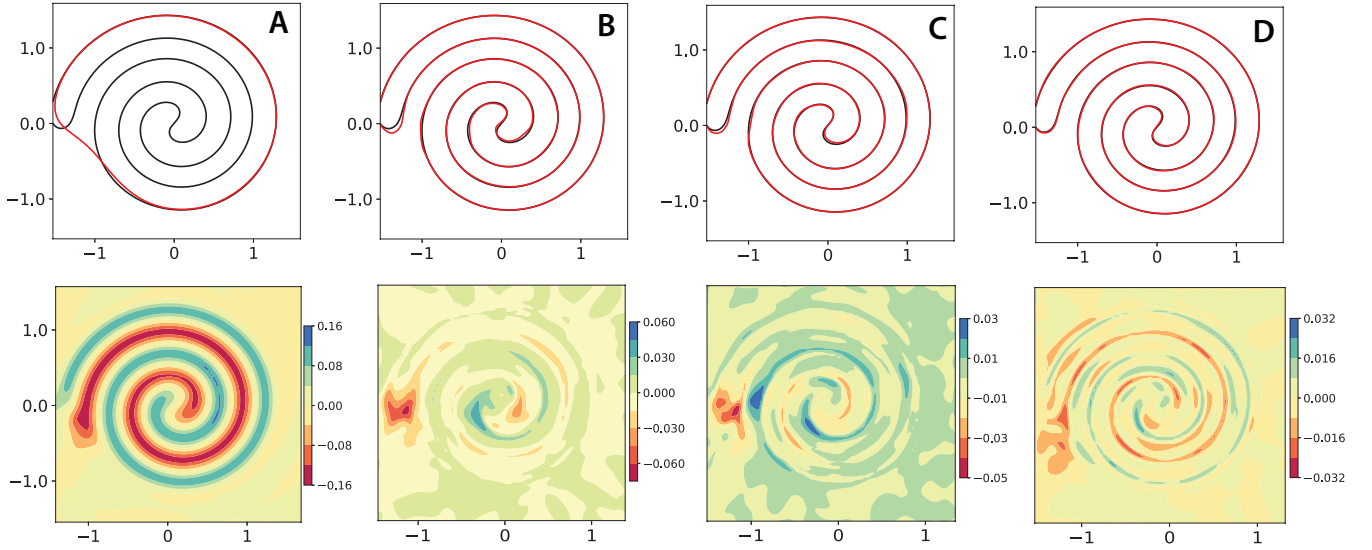}
\caption{\textbf{Solution snapshot of curvature-driven level-set problem}\footnotesize{ Top row) Zero contour of the level-set function visualized at simulation time $t=0.01s$ and after $2000$ epochs of training. Different columns correspond to different weighing strategies A) uniform weighing B) pinn-MGDA C) inverse-Dirichlet D) max/avg. Bottom row) The corresponding point-wise error of the PINN approximation of the level-set function in comparison to the ground truth solution.
}}
\label{fig:surface_contour}
\end{figure*}

\clearpage

\onecolumngrid
\section{Network initialization and parameterization}

\noindent Let $\mathcal{X}\in\mathbb{R}^{N\times d_{inp}}$ be the set of all input coordinates to train the neural network, where $N$ denotes the total number of data points and $d_{inp}$ the dimension of the input data, i.e. $d_{inp}=3$ corresponding to $(x,y,t)$. Moreover, $\bm{X}_\textrm{B}\in\mathbb{R}^{\vert\textrm{B}\vert\times d_{inp}}$ will denote a single batch, by randomly subsampling $\mathcal{X}$ into $N/\vert\textrm{B}\vert$ subsets. We then define an $n$-layer deep neural network with parameters $\bm{\theta}$ as a composite function $f_{\bm{\theta}}:\mathbb{R}^{\vert\textrm{B}\vert\times d_{inp}}\rightarrow\mathbb{R}^{\vert\textrm{B}\vert\times d_{out}}$, where $d_{out}$ corresponds to the dimension of the output variable. The output is usually computed as
\begin{equation}
	f_{\bm{\theta}}(\bm{X}_\textrm{B})=T_n(\sigma(T_{n-1}(\sigma(\dots\sigma(T_1(\bm{X}_\textrm{B})))))),
\end{equation}
with the affine linear maps $T_i(\bm{z}_{i-1})=\bm{z}_{i-1}\bm{W}^T_i + \bm{b}_i$, $i=1,\dots,n$, $\bm{z}_0=\bm{X}_\textrm{B}$, and output $\bm{Y}_\textrm{B}=\bm{z}_{n-1}\bm{W}^T_n + \bm{b}_n$. Here, $\bm{W}_i\in\mathbb{R}^{q_i\times q_{i-1}}$ describes the weights of the incoming edges of layer $i$ with $q_i$ neurons, $\bm{b}_i\in\mathbb{R}^{q_i}$ contains a scalar bias for each neuron and $\sigma(\cdot)$ is a non-linear activation function. The network parameters $\bm{W}_i, \bm{b}_i\subset\bm{\theta}$ are initialized using normal Xavier initialization \cite{glorot_understanding_2010} with $\bm{W}_i\sim\mathcal{G}(0,\Psi^2)$, where $\mathcal{G}$ is a Gaussian distribution with zero mean and standard deviation
\begin{equation}
    \Psi=g\cdot\sqrt{\frac{2}{q_i+ q_{i-1}}},
\end{equation}
and $\bm{b}_i=\bm{0}$. The gain $g$ was chosen to be $g=1$ for all activation functions except for $\tanh$ with a recommended gain of $g=\frac{5}{3}$. Before feeding $\bm{X}_\textrm{B}$ into the first layer of the neural network, we normalize it with respect to the training data $\mathcal{X}$ as
\begin{equation}
    \hat{\bm{X}}_\textrm{B}=\frac{\bm{X}_\textrm{B}-\bm{\mu}}{\bm{\psi}},
\end{equation}
where $\bm{\mu}=((\overline{\mathcal{X}_1},\dots,\overline{\mathcal{X}_{d_{inp}}}))$ and $\bm{\psi}=((\sqrt{\textrm{Var}\{\mathcal{X}_1\}},\dots, \sqrt{\textrm{Var}\{\mathcal{X}_{d_{inp}}\}}))$ are the column-wise mean and standard deviations of the training data $\mathcal{X}$.

\subsection*{Weights computation} \noindent We initalize all weights to  $\lambda_k=1,k= 1,\dots,K$. For all experiments we update the dynamic weights at the first batch of every fifth epoch. The moving average is then performed with $\alpha=0.5$. The mini-batches $\bm{X}_\textrm{B}$ are randomized after each epoch, such that for every update of the weights, $\lambda_k$ will be computed with respect to a different subset $\bm{X}_\textrm{B}$ of all training points $\mathcal{X}$. Furthermore, batching is only performed over the interior points $N_{\textrm{int}}$. As $N_{I},N_{B}\ll N_{\textrm{int}}$, the points for initial and boundary conditions don't require batching and remain fixed throughout the training.

\end{document}